\documentclass[sigplan,screen]{acmart}

\copyrightyear{2021}
\acmYear{2021}
\setcopyright{acmlicensed}
\acmConference[TyDe '21]{Proceedings of the 6th ACM SIGPLAN International Workshop on Type-Driven Development}{August 22, 2021}{Virtual Event, Republic of Korea}
\acmBooktitle{Proceedings of the 6th ACM SIGPLAN International Workshop on Type-Driven Development (TyDe '21), August 22, 2021, Virtual Event, Republic of Korea}
\acmPrice{15.00}
\acmDOI{10.1145/3471875.3472990}
\acmISBN{978-1-4503-8616-6/21/08}

\AtBeginDocument{%
  \providecommand\BibTeX{{%
    \normalfont B\kern-0.5em{\scshape i\kern-0.25em b}\kern-0.8em\TeX}}}

\usepackage{graphicx}

\usepackage{ifxetex,ifluatex}
\usepackage{fixltx2e} 
\usepackage{stmaryrd}
\usepackage{tikz}
\usepackage{mathtools}  
\usepackage{semantic}
\usepackage[]{agda}
\usepackage{fancyvrb}
\usepackage{stackrel}
\usepackage{tcolorbox}
\usepackage{todonotes}
\usepackage{todonotes, comment}

\usepackage{balance}
\usepackage{breakcites}
\DeclareUnicodeCharacter{393}{$\Gamma$}
\DeclareUnicodeCharacter{2081}{$_1$}
\DeclareUnicodeCharacter{2082}{$_2$}
\DeclareUnicodeCharacter{03BB}{$\lambda$}
\DeclareUnicodeCharacter{2200}{$\forall$}
\DeclareUnicodeCharacter{2192}{$\shortrightarrow$}
\DeclareUnicodeCharacter{2237}{$::$}
\DeclareUnicodeCharacter{03C3}{$\sigma$}
\DeclareUnicodeCharacter{03B1}{$\alpha$}
\DeclareUnicodeCharacter{27E6}{$\llbracket$}
\DeclareUnicodeCharacter{27E7}{$\rrbracket$}
\DeclareUnicodeCharacter{22A2}{$\vdash$}
\DeclareUnicodeCharacter{2236}{$:$}
\DeclareUnicodeCharacter{219D}{$\strans$}
\DeclareUnicodeCharacter{2208}{$\in$}
\DeclareUnicodeCharacter{2294}{$\sqcup$}
\DeclareUnicodeCharacter{2265}{$\geq$}
\DeclareUnicodeCharacter{228E}{$\uplus$}
\DeclareUnicodeCharacter{22A5}{$\bot$}
\DeclareUnicodeCharacter{22A4}{$\top$}
\DeclareUnicodeCharacter{225F}{$\stackrel{?}{=}$}
\DeclareUnicodeCharacter{209A}{$_p$}
\DeclareUnicodeCharacter{2287}{$\supseteq$}
\DeclareUnicodeCharacter{2080}{$_0$}
\DeclareUnicodeCharacter{2238}{$-$}
\DeclareUnicodeCharacter{2115}{$\mathbb{N}$}
\DeclareUnicodeCharacter{2209}{$\notin$}
\DeclareUnicodeCharacter{2261}{$\equiv$}

\def\minus{-}
\mathchardef\mplus="2B

\newcommand{\strans}{\leadsto}

\newcommand{\phalt}{\mathrm{halt}}

\usepackage{xcolor}
\definecolor{AgdaDatatype}  {HTML}{0000CD}
\definecolor{AgdaFunction}              {HTML}{006400}

\newcommand{\agdaData}[1]{\textcolor{AgdaDatatype}{\textsf #1}}
\newcommand{\agdaFunction}[1]{\textcolor{AgdaFunction}{\textsf #1}}

\begin{document}

\title{Actions You Can Handle: \\ Dependent Types for AI Plans}

\author{Alasdair Hill}
\affiliation{%
  \institution{Heriot-Watt University}
  \city{Edinburgh}
  \country{Scotland}}
\email{ath7@hw.ac.uk}

\author{Ekaterina Komendantskaya}
\affiliation{%
    \institution{Heriot-Watt University}
    \city{Edinburgh}
    \country{Scotland}
}
\email{e.komendantskaya@hw.ac.uk}

\author{Matthew L. Daggitt}
\affiliation{%
    \institution{Heriot-Watt University}
    \city{Edinburgh}
    \country{Scotland}
}
\email{m.daggitt@hw.ac.uk}

\author{Ronald P. A. Petrick }
\affiliation{%
    \institution{Heriot-Watt University}
    \city{Edinburgh}
    \country{Scotland}
}
\email{R.Petrick@hw.ac.uk}

\renewcommand{\shortauthors}{Hill et al.}

\begin{abstract}
  Verification of AI is a challenge that has engineering, algorithmic and programming language components.
 For example,  AI planners are deployed to model actions of autonomous agents. They comprise a number of searching algorithms that, given a set of specified properties, find a sequence of actions that satisfy these properties. Although AI planners are mature tools from the algorithmic and engineering points of view, they have limitations as programming languages. Decidable and efficient automated search entails restrictions on the syntax of the language, prohibiting use of higher-order properties or recursion. This paper proposes a methodology for embedding plans produced by AI planners into the dependently-typed language Agda, which enables users to reason about and verify more general and abstract properties of plans, and also provides a more holistic programming language infrastructure for modelling plan execution.   
\end{abstract}


\maketitle

\section{Introduction}\label{sec:intro}

Planning is a research area within AI that studies the automated generation of plans from
symbolic domain and problem specifications.
AI planners came into existence in the 1970s as an intersection between general problem solvers \cite{ernst1969gps},
situation calculus \cite{mccarthy1981some} and theorem proving~\cite{green1969theorem}.

Typically, the domain is represented by an abstract description of the \emph{world} and a set of actions that can be used to alter the world states (see Figure~\ref{fig:pddl-taxi-domain}). Planning problems in the domain consist of the \emph{initial state} of the world and a \emph{goal state} (see Figure~\ref{fig:pddl-taxi-problem}). The planner then produces a \emph{plan}, i.e. a sequence of actions, moving the world from the initial state to the goal state (see Figure~\ref{fig:pddl-taxi-solution}). In most domains, the plan produced must not only reach the goal state, but also satisfy other properties such as safety. These properties are encoded via the preconditions of actions. For example, a ``rotate'' action for a robotic arm might have the precondition that there are no obstacles in the way. The preconditions are taken into account by the planner when creating the plan, and therefore we shall refer to these  as \emph{intrinsic} properties.

Our previous work~\cite{SchwaabKHFPWH19,HillKP20} has shown that the operational and declarative semantics of AI planning can be abstractly specified by a simple calculus resembling Hoare Logic~\cite{hoare1969axiomatic}. Formalisation of this calculus in Agda~\cite{bove2009brief} allowed us to prove soundness of the operational semantics. Moreover,
in~\cite{HillKP20} we showed how the formalisation allowed us to semi-automatically verify that individual plans produced by AI planners are sound with respect to their formal semantics, and therefore that plans produced by the planner really do satisfy the desired intrinsic properties encoded in the action preconditions.

\subsection{Verifying extrinsic properties}\label{sec:extrinsic}

 In this paper, we extend this work to show that our Agda framework can be used to reason about plan properties that the planner itself either \emph{cannot} or \emph{should not} reason about. We will refer to these as \emph{extrinsic} properties.
 
 There are three main classes of extrinsic properties that we have identified:
\begin{enumerate}
\item \textbf{Inexpressible properties} - these are properties that cannot be expressed in the declarative specification language of the planner, for example because they involve high-order functions or unbounded state. A good example of such a property is that the plan produced is \emph{fair}. Fairness typically involves universally quantifying over all the agents in the problem and keeping track of and comparing state. As discussed in Section~\ref{sec:PDDL-expressivity}, such global properties are typically impossible to express as pre-conditions of individual actions in the baseline versions of planning languages such as PDDL~\cite{mcdermott1998pddl}. In contrast, it is easy to express to express and reason about them in a dependently-typed language like Agda.

\item \textbf{Unavailable properties} - these are properties whose evaluation requires world states that are not available at planning time. A good example of such a property is the fuel consumption of a robotic agent. Although the fuel used per action can be estimated at planning time, in practice the amount of the fuel required to carry out an action in the real-world may depend on real-time conditions such as weather, temperature or other local conditions. Therefore, even though it cannot be checked at planning time, it is still desirable to verify that during execution the robotic agent never starts an action that it has insufficient fuel to complete.

\item \textbf{Probable properties} - finally these are properties which plans produced by the planner have a high probability of satisfying. As an example of such a property, we once again consider fairness. Suppose our planner is assigning jobs to workers and we want to verify that the set of assignments does not exhibit gender bias. By default, if the planner does not have access to gender information you would expect the vast majority of plans to be fair. Nonetheless, it is possible that in certain circumstances some other part of the domain may act as a proxy for gender and result in plans that are biased. Such problems are widely known in data science and machine learning~\cite{o2016weapons}. Even if such a property can be added to the planning domain, the time complexity of planning algorithms is typically super-linear in the size of the domain. Therefore we argue that one should avoid encoding it in the problem domain and only verify the property holds of any produced plans. As the property failure rate is low, one can achieve significant speed-ups at planning time. 
\end{enumerate}

We view the distinction between intrinsic and extrinsic properties as mirroring the separation between the \emph{search} and \emph{control} components of AI modelling. The intrinsic properties are incorporated into PDDL domains and inform the PDDL search algorithms. In contrast, the extrinsic properties are imposed on the controller and our framework provides a methodology for guaranteeing that the controller does not violate them during the execution of the plan. We envisage this guarantee being used in one of two ways: firstly, prior to execution, the plan can be run in a simulated environment to check for extrinsic property violations. Secondly, during the live execution of the plan it can be used as a form of run-time checking that provides a \emph{formally verified} guarantee that the controller will never perform an action that violates the extrinsic properties. Note that in this paper we do not address what the controller should do in response to an imminent violation of an extrinsic properties. We discuss work on how one might provide feedback about the violation to the planner in Section~\ref{sec:relatedwork}.

\subsection{The technical approach}

The work presented in this paper builds upon the earlier work of~\cite{SchwaabKHFPWH19}, in which the Agda fomalisation of PDDL was first given. However, we substantially clarify and simplify that initial formalisation here. 

  Our novel technical contribution is the use of \emph{action handlers} as a means of integrating rich extrinsic properties expressed in the proof and programming environment of Agda with our previous PDDL formalisation. An action handler is a function that, given a state and an action, \emph{executes the action} by applying the action (seen as a function) to the state. The  handlers were introduced in \cite{SchwaabKHFPWH19} as an auxiliary means of establishing a correspondence between the declarative and the operational semantics of AI planning. 

  In this paper, action handlers become the central tool for building richer program and proof infrastructure around the plans produced by AI planners.
  In particular we use dependent-types to enrich the handlers with additional constraints representing extrinsic properties that should hold during plan execution.
  As a result, we obtain \emph{enriched action handlers} in which we can incorporate additional safety, security, fairness or other checks of arbitrary complexity which are then formally verified by Agda.
 Crucially, these extrinsic properties can be expressed and verified without altering either the native PDDL problem domain or its formal semantics. 
  Notably, the richer properties we seek to define and prove are specified at the type level.  From this point of view, this paper presents a non-trivial exercise in dependently-typed programming.

With regards to future applications, this paper can be seen as a prototype for embedding existing automated reasoning tools within dependently-typed modelling environments.
  For example, we can perform higher-order reasoning (in Agda's interactive style) on top of the first-order proof search already performed by the AI planner.  This substantially extends the modelling power of the AI planners, as in Agda we can encode many properties that PDDL cannot. This includes function definitions, universal and existential quantification, action dependencies and higher-order quantification. 
  We argue that this approach promises to play an important role in verification of complex AI applications.



\subsection{Road map}

We proceed as follows. Section~\ref{sec:ah} contains a brief summary of the PDDL language that is used for planners, illustrated by using a classic taxi planning problem. We then recap the Agda formalisation of plans first developed in~\cite{SchwaabKHFPWH19, HillKP20}, including the notion of the canonical action handler as motivated by the running example. Section~\ref{sec:policies} introduces the novel method of enriched handlers by illustrating how to model and incorporate rich extrinsic verification properties into the type level of handlers. Section~\ref{sec:future} discusses future work, including on the handling of failure of the extrinsic properties and how this work relates to Explainable AI.

\section{PPDL, Plans, Action Handlers \& Agda}
\label{sec:ah}

In this section, we provide an introduction to the PDDL planning language and the essential parts of the Agda formalisation accompanying~\cite{HillKP20}, thereby providing some clarification and simplification of that formalisation. 
This will then pave the way to Section~\ref{sec:policies} in which we explain how to extend the formalisation to allow the embedding of extrinsic properties. We refer the reader directly to~\cite{HillKP20} for more theoretical aspects of the previous work. 

\begin{figure}[t!]
    \centering
    \hfill\break
  \begin{Verbatim}[frame=single,rulecolor=\color{black},label=$\backslash \backslash$ 1. The notion of domain]
(define (domain taxi) 
  (:requirements :strips :typing)
  \end{Verbatim}    
  \hfill

  \begin{Verbatim}[frame=single,rulecolor=\color{black},label=$\backslash \backslash$ 2. Types]
  (:types taxi location person)
  \end{Verbatim}  
  \hfill

  \begin{Verbatim}[frame=single,rulecolor=\color{black},label=$\backslash \backslash$ 3. Predicates]
  (:predicates   
   (taxiIn ?obj1 - taxi ?l1 - location)
   (personIn ?obj1 - person ?l1 - location))
  \end{Verbatim}  
  \hfill

\begin{Verbatim}[frame=single,rulecolor=\color{black},label=$\backslash \backslash$ 4. Actions] 
  (:action drive_passenger
    :parameters
      (?t1 - taxi ?p1 - person 
       ?l1 - location ?l2 - location)
\end{Verbatim}
\hfill

\begin{Verbatim}[frame=single,rulecolor=\color{black},label=$\backslash \backslash$ 5. Action preconditions and effects]
    :precondition 
      (and
        (taxiIn ?t1 ?l1)
        (personIn ?p1 ?l1)) 
    :effect
      (and
        (not (taxiIn ?t1 ?l1))
        (not (personIn ?p1 ?l1))
        (taxiIn ?t1 ?l2)
        (personIn ?p1 ?l2)))
\end{Verbatim}
\begin{Verbatim}
  (:action drive
    :parameters
      (?t1 - taxi ?l1 - location ?l2 - location)
    :precondition
      (taxiIn ?t1 ?l1)
    :effect
      (and
        (not (taxiIn ?t1 ?l1))
        (taxiIn ?t1 ?l2))))
\end{Verbatim}

      \caption{\small{\emph{The PDDL Taxi Domain, with main logical blocks outlined in boxes.}}}
    \label{fig:pddl-taxi-domain}
\end{figure}

\begin{figure}[t!]
   \centering
        \begin{verbatim}
(define (problem taxi)
  (:domain taxi)
  (:objects
      taxi1 taxi2 taxi3 - taxi
      person1 person2 person3 - person
      loc1 loc2 loc3 - location)
  (:init  (taxiIn taxi1 loc1)
          (taxiIn taxi2 loc2)
          (taxiIn taxi3 loc3)
          (personIn person1 loc1)
          (personIn person2 loc2)
          (personIn person3 loc3))
  (:goal (and  (taxiIn taxi1 loc2)
               (personIn person1 loc3)
               (personIn person3 loc1))))
    \end{verbatim}
    \caption{\emph{\small{A Taxi planning problem expressed in \emph{PDDL}. Initial state: There are three taxis with taxi1 being in loc1, taxi2 in loc2 and taxi3 in loc3. There are also three people with person1 being in loc1, person2 in loc2 and person3 in loc3. Goal state: taxi1 is in loc2, person1 is in loc3 and person3 is in loc1.
       } }}
    \label{fig:pddl-taxi-problem}
\end{figure}

\begin{figure}
\begin{Verbatim}
plan =
  (drive_passenger taxi3 person3 loc3 loc1);
  (drive taxi1 loc1 loc2);
  (drive_passenger taxi3 person1 loc1 loc3)
\end{Verbatim}
\caption{One possible solution to the Taxi planning problem in Figure~\ref{fig:pddl-taxi-problem}}
\label{fig:pddl-taxi-solution}
\end{figure}

\subsection{PDDL Syntax}

Many versions of planning languages were proposed, and
 the \emph{Planning Domain and Definition Language (PDDL)}~\cite{mcdermott1998pddl} aimed to standardise them.
One notable design decision of PDDL is the splitting of the planning problem into
\emph{domain} and \emph{problem} descriptions. The domain describes the predicates and admissible actions (as shown in Figure~\ref{fig:pddl-taxi-domain}), while
the problem description defines specific \emph{initial} and \emph{goal} states (Figure~\ref{fig:pddl-taxi-problem}).

We begin by explaining how each part of a planning domain and a planning problem are translated from PDDL into our dependently typed framework. In general, we maintain two kinds of Agda files. The first are the files which hold general definitions of the PDDL syntax, contexts and inference rules and are parametrised by an abstract domain. The second are example files which contain concrete encodings of the planning example in question, which then instantiate the generic modules with the corresponding parts of the encoding. The user only has to interact with the latter to specify the properties over their PDDL domain, whilst former can be hidden from their view.

\paragraph{An abstract planning domain}

Types, predicates and actions (blocks 2, 3 \& 4 in Figure~\ref{fig:pddl-taxi-domain}) are the basic components of any PDDL domain definition, and abstractly these are represented as three Agda sets \AgdaField{Type}, \AgdaField{Predicate} and \AgdaField{Action}.

To indicate whether a predicate is true or false we map it to a \agdaData{Polarity}, a set that contains two elements, $+$ and $-$. We then have a notion of state that is a list of polarities mapped to predicates. The \agdaData{State} type is used to represent preconditions and effects of actions in a domain as well as the goal state in the problem specification. We will refer to specific instances of the \agdaData{State} type using the types \agdaData{Precondition}, \agdaData{Effect} and \agdaData{GoalState}.

\begin{code}
\>[0]\AgdaFunction{State}\AgdaSpace{}%
\AgdaSymbol{:}\AgdaSpace{}%
\AgdaPrimitiveType{Set}\<%
\\
\>[0]\AgdaFunction{State}\AgdaSpace{}%
\AgdaSymbol{=}\AgdaSpace{}%
\AgdaDatatype{List}\AgdaSpace{}%
\AgdaFunction{(Polarity × Predicate)}\<%
\end{code}

The notion of actions' preconditions and effects (block 5 in Figure~\ref{fig:pddl-taxi-domain}) are defined generically as the \AgdaRecord{ActionDescription} record:
\begin{code}
\>[0]\AgdaKeyword{record}\AgdaSpace{}%
\AgdaRecord{ActionDescription}\AgdaSpace{}%
\AgdaSymbol{:}\AgdaSpace{}%
\AgdaPrimitiveType{Set}\AgdaSpace{}%
\AgdaKeyword{where}\<%
\\
\>[0][@{}l@{\AgdaIndent{0}}]%
\>[2]\AgdaKeyword{field}\<%
\\
\>[2][@{}l@{\AgdaIndent{0}}]%
\>[4]\AgdaField{preconditions}\AgdaSpace{}%
\>[10]\AgdaSymbol{:}\AgdaSpace{}%
\AgdaFunction{Preconditions}\<%
\\
\>[4]\AgdaField{effects}\AgdaSpace{}%
\>[10]\AgdaSymbol{:}\AgdaSpace{}%
\AgdaFunction{Effects}\<%
\end{code}
and a context maps every \AgdaField{Action} to an \AgdaRecord{ActionDescription}:
\begin{code}
\>[0]\AgdaFunction{Context}\AgdaSpace{}%
\AgdaSymbol{:}\AgdaSpace{}%
\AgdaPrimitiveType{Set}\<%
\\
\>[0]\AgdaFunction{Context}\AgdaSpace{}%
\AgdaSymbol{=}\AgdaSpace{}%
\AgdaField{Action}\AgdaSpace{}%
\AgdaSymbol{→}\AgdaSpace{}%
\AgdaRecord{ActionDescription}\<%
\end{code}
This then allows us to represent an abstract planning domain (block 1 in Figure~\ref{fig:pddl-taxi-domain}) as the following record:
\begin{code}
\>[0]\AgdaKeyword{record}\AgdaSpace{}%
\AgdaRecord{Domain}\AgdaSpace{}%
\AgdaSymbol{:}\AgdaSpace{}%
\AgdaPrimitiveType{Set₁}\AgdaSpace{}%
\AgdaKeyword{where}\<%
\\
\>[0][@{}l@{\AgdaIndent{0}}]%
\>[2]\AgdaKeyword{field}\<%
\\
\>[2][@{}l@{\AgdaIndent{0}}]%
\>[4]\AgdaField{Type}\AgdaSpace{}%
\>[10]\AgdaSymbol{:}\AgdaSpace{}%
\AgdaPrimitiveType{Set}\<%
\\
\>[4]\AgdaField{Action}\AgdaSpace{}%
\>[10]\AgdaSymbol{:}\AgdaSpace{}%
\AgdaPrimitiveType{Set}\<%
\\
\>[4]\AgdaField{Predicate}\AgdaSpace{}%
\>[10]\AgdaSymbol{:}\AgdaSpace{}%
\AgdaPrimitiveType{Set}\<%
\\
\>[4]\AgdaField{Γ}\AgdaSpace{}%
\>[10]\AgdaSymbol{:}\AgdaSpace{}%
\AgdaFunction{Context}
\\
\>[4]\AgdaOperator{\AgdaField{\AgdaUnderscore{}≟ₚ\AgdaUnderscore{}}}\AgdaSpace{}%
\>[10]\AgdaSymbol{:}\AgdaSpace{}%
\AgdaFunction{DecidableEquality}\AgdaSpace{}%
\AgdaField{Predicate}\<%
\end{code}

\paragraph{The taxi planning domain}

We can then instantiate the taxi domain as follows. To describe PDDL types one simply needs to create a data type in Agda with the required types as constructors.
\begin{code}
\>[0]\AgdaKeyword{data}\AgdaSpace{}%
\AgdaDatatype{Type}\AgdaSpace{}%
\AgdaSymbol{:}\AgdaSpace{}%
\AgdaPrimitiveType{Set}\AgdaSpace{}%
\AgdaKeyword{where}\<%
\\
\>[0][@{}l@{\AgdaIndent{0}}]%
\>[2]\AgdaInductiveConstructor{taxi}\AgdaSpace{}%
\AgdaInductiveConstructor{location}\AgdaSpace{}%
\AgdaInductiveConstructor{person}\AgdaSpace{}%
\AgdaSymbol{:}\AgdaSpace{}%
\AgdaDatatype{Type}\<%
\end{code}
In PDDL, objects are first introduced implicitly as typed variables within the block that defines predicates, and only later does the planning problem give an explicit set of objects that can be used to instantiate these variables. For convenience, in our formalisation we combine these two separate notions of objects and variables into a single \agdaData{Object} data type whose constructors are indexed by \agdaData{Type}s. The number of objects for each constructor are given by a finite number indicated by \agdaData{Fin}. For example if \agdaFunction{numberOfTaxis} was equal to 3 then we can construct taxis: \agdaFunction{taxi 0, taxi 1, taxi 2}. Thus, the second block of Figure~\ref{fig:pddl-taxi-domain} boils down to the following data declarations: 
\begin{code}
\>[0]\AgdaKeyword{data}\AgdaSpace{}%
\AgdaDatatype{Object}\AgdaSpace{}%
\AgdaSymbol{:}\AgdaSpace{}%
\AgdaDatatype{Type}\AgdaSpace{}%
\AgdaSymbol{->}\AgdaSpace{}%
\AgdaPrimitiveType{Set}\AgdaSpace{}%
\AgdaKeyword{where}\<%
\\
\>[0][@{}l@{\AgdaIndent{0}}]%
\>[2]\AgdaInductiveConstructor{taxi}\AgdaSpace{}%
\>[10]\AgdaSymbol{:}\AgdaSpace{}%
\AgdaDatatype{Fin}\AgdaSpace{}%
\AgdaFunction{numberOfTaxis}\AgdaSpace{}%
\>[20]\AgdaSymbol{→}\AgdaSpace{}%
\AgdaDatatype{Object}\AgdaSpace{}%
\AgdaInductiveConstructor{taxi}\<%
\\
\>[2]\AgdaInductiveConstructor{location}\AgdaSpace{}%
\>[10]\AgdaSymbol{:}\AgdaSpace{}%
\AgdaDatatype{Fin}\AgdaSpace{}%
\AgdaFunction{numberOfLocations}\AgdaSpace{}%
\>[20]\AgdaSymbol{→}\AgdaSpace{}%
\AgdaDatatype{Object}\AgdaSpace{}%
\AgdaInductiveConstructor{location}\<%
\\
\>[2]\AgdaInductiveConstructor{person}\AgdaSpace{}%
\>[10]\AgdaSymbol{:}\AgdaSpace{}%
\AgdaDatatype{Fin}\AgdaSpace{}%
\AgdaFunction{numberOfPeople}\AgdaSpace{}%
\>[20]\AgdaSymbol{→}\AgdaSpace{}%
\AgdaDatatype{Object}\AgdaSpace{}%
\AgdaInductiveConstructor{person}\<%
\end{code}
In practice, developing this prototype system further may require one to properly distinguish between objects and variables, but we leave this for future work.  
In \cite{SchwaabKHFPWH19} objects were defined simply by naming them as constructors. However, in this paper we need to add additional information about the number of available taxis in order to reason about extrinsic properties of plans, such as fairness.

We can now define predicates over typed objects, closely mimicking the PDDL syntax in block 3 of Figure~\ref{fig:pddl-taxi-domain}.    
\begin{code}
\>[0]\AgdaKeyword{data}\AgdaSpace{}%
\AgdaDatatype{Predicate}\AgdaSpace{}%
\AgdaSymbol{:}\AgdaSpace{}%
\AgdaPrimitiveType{Set}\AgdaSpace{}%
\AgdaKeyword{where}\<%
\\
\>[0][@{}l@{\AgdaIndent{0}}]%
\>[2]\AgdaInductiveConstructor{taxiIn}\AgdaSpace{}%
\>[10]\AgdaSymbol{:}\AgdaSpace{}%
\AgdaDatatype{Object}\AgdaSpace{}%
\AgdaInductiveConstructor{taxi}\AgdaSpace{}%
\>[20]\AgdaSymbol{→}\AgdaSpace{}%
\AgdaDatatype{Object}\AgdaSpace{}%
\AgdaInductiveConstructor{location}\AgdaSpace{}%
\>[30]\AgdaSymbol{→}\AgdaSpace{}%
\AgdaDatatype{Predicate}\<%
\\
\>[2]\AgdaInductiveConstructor{personIn}\AgdaSpace{}%
\>[10]\AgdaSymbol{:}\AgdaSpace{}%
\AgdaDatatype{Object}\AgdaSpace{}%
\AgdaInductiveConstructor{person}\AgdaSpace{}%
\>[20]\AgdaSymbol{→}\AgdaSpace{}%
\AgdaDatatype{Object}\AgdaSpace{}%
\AgdaInductiveConstructor{location}\AgdaSpace{}%
\>[30]\AgdaSymbol{→}\AgdaSpace{}%
\AgdaDatatype{Predicate}\<%
\end{code}
Actions (see block $3$, $4$ and $5$ in Figure~\ref{fig:pddl-taxi-domain}) are defined as another data type. 
\begin{code}
\>[0]\AgdaKeyword{data}\AgdaSpace{}%
\AgdaDatatype{Action}\AgdaSpace{}%
\AgdaSymbol{:}\AgdaSpace{}%
\AgdaPrimitiveType{Set}\AgdaSpace{}%
\AgdaKeyword{where}\<%
\\
\>[0][@{}l@{\AgdaIndent{0}}]%
\>[2]\AgdaInductiveConstructor{drive}\AgdaSpace{}%
\AgdaSymbol{:}\AgdaSpace{}%
\AgdaDatatype{Object}\AgdaSpace{}%
\AgdaInductiveConstructor{taxi}%
\>[82I]\AgdaSymbol{→}\AgdaSpace{}%
\AgdaDatatype{Object}\AgdaSpace{}%
\AgdaInductiveConstructor{location}\<%
\\
\>[.][@{}l@{}]\<[82I]%
\>[22]\AgdaSymbol{→}\AgdaSpace{}%
\AgdaDatatype{Object}\AgdaSpace{}%
\AgdaInductiveConstructor{location}\<%
\\
\>[22]\AgdaSymbol{→}\AgdaSpace{}%
\AgdaDatatype{Action}\<%
\\
\>[0][@{}l@{\AgdaIndent{0}}]%
\>[2]\AgdaInductiveConstructor{drivePassenger}\AgdaSpace{}%
\AgdaSymbol{:}\AgdaSpace{}%
\AgdaComment{... }\<
\end{code}
\noindent The context that details each action's preconditions and effects can be easily instantiated in a manner that is close to the PDDL syntax:

\begin{code}
\>[0]\AgdaFunction{$\Gamma$}\AgdaSpace{}%
\AgdaSymbol{:}\AgdaSpace{}%
\AgdaFunction{Context}\<%
\\
\>[0]\AgdaFunction{$\Gamma$}\AgdaSpace{}%
\AgdaSymbol{(}\AgdaInductiveConstructor{drive}\AgdaSpace{}%
\AgdaBound{t1}\AgdaSpace{}%
\AgdaBound{l1}\AgdaSpace{}%
\AgdaBound{l2}\AgdaSymbol{)}\AgdaSpace{}%
\AgdaSymbol{=}\<%
\\
\>[0][@{}l@{\AgdaIndent{0}}]%
\>[2]\AgdaKeyword{record}\AgdaSpace{}%
\AgdaSymbol{\{}\<%
\\
\>[2][@{}l@{\AgdaIndent{0}}]%
\>[4]\AgdaField{preconditions}\AgdaSpace{}%
\AgdaSymbol{=}\<%
\\
\>[4][@{}l@{\AgdaIndent{0}}]%
\>[6]\AgdaSymbol{(}\AgdaInductiveConstructor{+}\AgdaSpace{}%
\AgdaOperator{\AgdaInductiveConstructor{,}}\AgdaSpace{}%
\AgdaInductiveConstructor{taxiIn}\AgdaSpace{}%
\AgdaBound{t1}\AgdaSpace{}%
\AgdaBound{l1}\AgdaSymbol{)}\AgdaSpace{}%
\AgdaOperator{\AgdaInductiveConstructor{∷}}\AgdaSpace{}%
\AgdaInductiveConstructor{[]}\AgdaSpace{}%
\AgdaSymbol{;}\<%
\\
\>[4]\AgdaField{effects}\AgdaSpace{}%
\AgdaSymbol{=}\<%
\\
\>[4][@{}l@{\AgdaIndent{0}}]%
\>[6]\AgdaSymbol{(}\AgdaInductiveConstructor{-}\AgdaSpace{}%
\AgdaOperator{\AgdaInductiveConstructor{,}}\AgdaSpace{}%
\AgdaInductiveConstructor{taxiIn}\AgdaSpace{}%
\AgdaBound{t1}\AgdaSpace{}%
\AgdaBound{l1}\AgdaSymbol{)}\AgdaSpace{}%
\AgdaOperator{\AgdaInductiveConstructor{∷}}\<%
\\
\>[6]\AgdaSymbol{(}\AgdaInductiveConstructor{+}\AgdaSpace{}%
\AgdaOperator{\AgdaInductiveConstructor{,}}\AgdaSpace{}%
\AgdaInductiveConstructor{taxiIn}\AgdaSpace{}%
\AgdaBound{t1}\AgdaSpace{}%
\AgdaBound{l2}\AgdaSymbol{)}\AgdaSpace{}%
\AgdaOperator{\AgdaInductiveConstructor{∷}}\AgdaSpace{}%
\AgdaInductiveConstructor{[]}\AgdaSpace{}%
\AgdaSymbol{\}}\<%
\\
\>[0]\AgdaComment{... }\<
\end{code}




\paragraph{The planning problem}

The specific planning problem (see Figure~\ref{fig:pddl-taxi-problem}) needs to be defined concretely, by providing an initial \agdaData{World} and a \agdaData{GoalState}. Intuitively, a \agdaData{World} is a logical description of the predicates that are true. We operate under the same \emph{closed world} assumption as PDDL where all predicates that are not contained in a world are assumed to be false. 
\begin{code}
\>[0]\AgdaFunction{World}\AgdaSpace{}%
\AgdaSymbol{:}\AgdaSpace{}%
\AgdaPrimitiveType{Set}\<%
\\
\>[0]\AgdaFunction{World}\AgdaSpace{}%
\AgdaSymbol{=}\AgdaSpace{}%
\AgdaDatatype{List}\AgdaSpace{}%
\AgdaField{Predicate}\<%
\end{code}
Whereas we use a \agdaData{World} to represent the initial state, we need to be able to talk about specific negative predicates in the \agdaData{GoalState} so it is a simple alias for \agdaData{State}.

\begin{code}
\>[0]\AgdaFunction{initialWorld}\AgdaSpace{}%
\AgdaSymbol{:}\AgdaSpace{}%
\AgdaFunction{World}\<%
\\
\>[0]\AgdaFunction{initialWorld}\AgdaSpace{}%
\AgdaSymbol{=}\<%
\\
\>[0][@{}l@{\AgdaIndent{0}}]%
\>[2]\AgdaInductiveConstructor{taxiIn}\AgdaSpace{}%
\AgdaSymbol{(}\AgdaInductiveConstructor{taxi}\AgdaSpace{}%
\AgdaNumber{0}\AgdaSymbol{)}\AgdaSpace{}%
\AgdaSymbol{(}\AgdaInductiveConstructor{location}\AgdaSpace{}%
\AgdaNumber{0}\AgdaSymbol{)}\AgdaSpace{}%
\AgdaOperator{\AgdaInductiveConstructor{∷}}\<%
\\
\>[2]\AgdaInductiveConstructor{taxiIn}\AgdaSpace{}%
\AgdaSymbol{(}\AgdaInductiveConstructor{taxi}\AgdaSpace{}%
\AgdaNumber{1}\AgdaSymbol{)}\AgdaSpace{}%
\AgdaSymbol{(}\AgdaInductiveConstructor{location}\AgdaSpace{}%
\AgdaNumber{1}\AgdaSymbol{)}\AgdaSpace{}%
\AgdaOperator{\AgdaInductiveConstructor{∷}}\<%
\\
\>[2]\AgdaInductiveConstructor{taxiIn}\AgdaSpace{}%
\AgdaSymbol{(}\AgdaInductiveConstructor{taxi}\AgdaSpace{}%
\AgdaNumber{2}\AgdaSymbol{)}\AgdaSpace{}%
\AgdaSymbol{(}\AgdaInductiveConstructor{location}\AgdaSpace{}%
\AgdaNumber{2}\AgdaSymbol{)}\AgdaSpace{}%
\AgdaOperator{\AgdaInductiveConstructor{∷}}\<%
\\
\>[2]\AgdaInductiveConstructor{personIn}\AgdaSpace{}%
\AgdaSymbol{(}\AgdaInductiveConstructor{person}\AgdaSpace{}%
\AgdaNumber{0}\AgdaSymbol{)}\AgdaSpace{}%
\AgdaSymbol{(}\AgdaInductiveConstructor{location}\AgdaSpace{}%
\AgdaNumber{0}\AgdaSymbol{)}\AgdaSpace{}%
\AgdaOperator{\AgdaInductiveConstructor{∷}}\<%
\\
\>[2]\AgdaInductiveConstructor{personIn}\AgdaSpace{}%
\AgdaSymbol{(}\AgdaInductiveConstructor{person}\AgdaSpace{}%
\AgdaNumber{1}\AgdaSymbol{)}\AgdaSpace{}%
\AgdaSymbol{(}\AgdaInductiveConstructor{location}\AgdaSpace{}%
\AgdaNumber{1}\AgdaSymbol{)}\AgdaSpace{}%
\AgdaOperator{\AgdaInductiveConstructor{∷}}\<%
\\
\>[2]\AgdaInductiveConstructor{personIn}\AgdaSpace{}%
\AgdaSymbol{(}\AgdaInductiveConstructor{person}\AgdaSpace{}%
\AgdaNumber{2}\AgdaSymbol{)}\AgdaSpace{}%
\AgdaSymbol{(}\AgdaInductiveConstructor{location}\AgdaSpace{}%
\AgdaNumber{2}\AgdaSymbol{)}\AgdaSpace{}%
\AgdaOperator{\AgdaInductiveConstructor{∷}}\<%
\\
\>[2]\AgdaInductiveConstructor{[]}\<%
\end{code}
\begin{code}
\\[\AgdaEmptyExtraSkip]%
\>[0]\AgdaFunction{goalState}\AgdaSpace{}%
\AgdaSymbol{:}\AgdaSpace{}%
\AgdaFunction{GoalState}\<%
\\
\>[0]\AgdaFunction{goalState}\AgdaSpace{}%
\AgdaSymbol{=}\<%
\\
\>[0][@{}l@{\AgdaIndent{0}}]%
\>[2]\AgdaSymbol{(}\AgdaInductiveConstructor{+}\AgdaSpace{}%
\AgdaOperator{\AgdaInductiveConstructor{,}}\AgdaSpace{}%
\AgdaInductiveConstructor{taxiIn}\AgdaSpace{}%
\AgdaSymbol{(}\AgdaInductiveConstructor{taxi}\AgdaSpace{}%
\AgdaNumber{0}\AgdaSymbol{)}\AgdaSpace{}%
\AgdaSymbol{(}\AgdaInductiveConstructor{location}\AgdaSpace{}%
\AgdaNumber{1}\AgdaSymbol{))}\AgdaSpace{}%
\AgdaOperator{\AgdaInductiveConstructor{∷}}\<%
\\
\>[2]\AgdaSymbol{(}\AgdaInductiveConstructor{+}\AgdaSpace{}%
\AgdaOperator{\AgdaInductiveConstructor{,}}\AgdaSpace{}%
\AgdaInductiveConstructor{personIn}\AgdaSpace{}%
\AgdaSymbol{(}\AgdaInductiveConstructor{person}\AgdaSpace{}%
\AgdaNumber{0}\AgdaSymbol{)}\AgdaSpace{}%
\AgdaSymbol{(}\AgdaInductiveConstructor{location}\AgdaSpace{}%
\AgdaNumber{2}\AgdaSymbol{))}\AgdaSpace{}%
\AgdaOperator{\AgdaInductiveConstructor{∷}}\<%
\\
\>[2]\AgdaSymbol{(}\AgdaInductiveConstructor{+}\AgdaSpace{}%
\AgdaOperator{\AgdaInductiveConstructor{,}}\AgdaSpace{}%
\AgdaInductiveConstructor{personIn}\AgdaSpace{}%
\AgdaSymbol{(}\AgdaInductiveConstructor{person}\AgdaSpace{}%
\AgdaNumber{2}\AgdaSymbol{)}\AgdaSpace{}%
\AgdaSymbol{(}\AgdaInductiveConstructor{location}\AgdaSpace{}%
\AgdaNumber{0}\AgdaSymbol{))}\AgdaSpace{}%
\AgdaOperator{\AgdaInductiveConstructor{∷}}\<%
\\
\>[2]\AgdaInductiveConstructor{[]}\<%
\end{code}

\paragraph{Plans}

One of the most popular early planners was the Stanford Research Institute Problem Solver (STRIPS)~\cite{strips}
which was created to address the problems faced by a robot in rearranging objects and in navigating. The STRIPS planner will perform an automatic search for a plan that moves from the initial world to the goal state  defined in the domain. One such plan that it might find for the problem outlined so far is shown in Figure~\ref{fig:pddl-taxi-solution}.


We define a \agdaData{Plan} as a list of actions, (renaming the empty list to \AgdaInductiveConstructor{halt} to improve readability).
\begin{code}
\>[0]\AgdaFunction{Plan}\AgdaSpace{}%
\AgdaSymbol{:}\AgdaSpace{}%
\AgdaPrimitiveType{Set}\<%
\\
\>[0]\AgdaFunction{Plan}\AgdaSpace{}%
\AgdaSymbol{=}\AgdaSpace{}%
\AgdaDatatype{List}\AgdaSpace{}%
\AgdaFunction{Action}\<%
\end{code}
The plan shown in Figure~\ref{fig:pddl-taxi-solution} can then be defined as:
\begin{code}
\>[0]\AgdaFunction{plan}\AgdaSpace{}%
\AgdaSymbol{:}\AgdaSpace{}%
\AgdaDatatype{Plan}\<%
\\
\>[0]\AgdaFunction{plan}\AgdaSpace{}%
\AgdaSymbol{=}%
\>[220I]\AgdaSymbol{(}\AgdaInductiveConstructor{drive}\AgdaSpace{}%
\AgdaInductiveConstructor{taxi1}\AgdaSpace{}%
\AgdaInductiveConstructor{loc1}\AgdaSpace{}%
\AgdaInductiveConstructor{loc2}\AgdaSymbol{)}\AgdaSpace{}%
\AgdaOperator{\AgdaInductiveConstructor{∷}}\<%
\\
\>[.][@{}l@{}]\<[220I]%
\>[8]\AgdaSymbol{(}\AgdaInductiveConstructor{drivePassenger}\AgdaSpace{}%
\AgdaInductiveConstructor{taxi3}\AgdaSpace{}%
\AgdaInductiveConstructor{person3}\AgdaSpace{}%
\AgdaInductiveConstructor{loc3}\AgdaSpace{}%
\AgdaInductiveConstructor{loc1}\AgdaSymbol{)}\AgdaSpace{}%
\AgdaOperator{\AgdaInductiveConstructor{∷}}\<%
\\
\>[8]\AgdaSymbol{(}\AgdaInductiveConstructor{drivePassenger}\AgdaSpace{}%
\AgdaInductiveConstructor{taxi3}\AgdaSpace{}%
\AgdaInductiveConstructor{person1}\AgdaSpace{}%
\AgdaInductiveConstructor{loc1}\AgdaSpace{}%
\AgdaInductiveConstructor{loc3}\AgdaSymbol{)}\AgdaSpace{}%
\AgdaOperator{\AgdaInductiveConstructor{∷}}\<%
\\
\>[8]\AgdaInductiveConstructor{halt}\<%
\end{code}

These are the main building blocks that we expect to receive from the given AI planner. 

\subsection{Expressivity of PDDL}
\label{sec:PDDL-expressivity}

PDDL is a very expressive language with many extensions. PDDL 1.2 usually operates under a closed world assumption and expresses domains using the STRIPS assumption where actions effects are applied by adding and deleting predicates to a given world.
The closed world requirement implies the use of first-order logic without function symbols (which guarantees finite domains when defining the models).
The problem with functions, especially with recursive functions, is that they can make domains infinite. For example, it only takes one nullary and one unary function to generate the set of natural numbers. 

PDDL 1.2 also allows for the expression of types with type hierarchy, equalities over objects, existential and universal quantification over preconditions and conditional effects. Conditional effects are effects that will only be applied when a list of preconditions hold true. In PDDL 2.1 there is also a definition of \emph{numeric fluents} that allow for the reasoning about numbers such as comparing and adding numbers.
PDDL 2.1 also introduces negative preconditions and durative actions. Durative actions add the concept of time to actions. Finally PDDL 3 adds strong and soft constraints that can be applied across a planning problem. Strong constraints can allow for the statement of certain implications to hold across every state during the execution of a plan. Soft constraints, also known as preferences, introduce soft goals that a user would prefer a planner to satisfy but are not necessary to satisfy for a valid plan. In this paper we will mainly focus on a subset of PDDL 1.2 under the closed world assumption.

Two of the above restrictions in particular are the subject of the syntactic (type-driven)  extensions we propose in this paper:
we do rely on arbitrary functions in our development, and we open ways to surpass the closed world assumption, by embedding the plans in a wider programming and modelling environment.  We also use higher-order functions and predicates to express some more sophisticated properties, for example calculating the number of taxi's that satisfy a certain property as discussed in Section~\ref{sec:fairness-example}. 

\subsection{Validating intrinsic properties of a plan}


We now briefly overview the calculus in which Agda validates the plans given by AI planners.  The calculus is very simple, consisting of just two rules: action sequencing and halting.
The main intuition behind the rules is that the goal state as well as the actions defined in $\Gamma$ describe
\emph{minimal} preconditions and effects, whereas plans are executed on potentially larger states.

We define a satisfaction relation between a world $w$ and a state $S$, denoted $w {\in} \langle S \rangle$, where a world satisfies a state when all positively mapped predicates in a state are in the world and all negatively mapped predicates are not in the world.
\begin{code}
\>[0]\AgdaOperator{\AgdaFunction{\AgdaUnderscore{}∈⟨\AgdaUnderscore{}⟩}}\AgdaSpace{}%
\AgdaSymbol{:}\AgdaSpace{}%
\AgdaFunction{World}\AgdaSpace{}%
\AgdaSymbol{→}\AgdaSpace{}%
\AgdaFunction{State}\AgdaSpace{}%
\AgdaSymbol{→}\AgdaSpace{}%
\AgdaPrimitiveType{Set}\<%
\\
\>[0]\AgdaBound{w}\AgdaSpace{}%
\AgdaOperator{\AgdaFunction{∈⟨}}\AgdaSpace{}%
\AgdaBound{S}\AgdaSpace{}%
\AgdaOperator{\AgdaFunction{⟩}}\AgdaSpace{}%
\AgdaSymbol{=}%
\>[30I]\AgdaSymbol{(∀}\AgdaSpace{}%
\AgdaBound{a}\AgdaSpace{}%
\AgdaSymbol{→}\AgdaSpace{}%
\AgdaSymbol{(}\AgdaInductiveConstructor{+}\AgdaSpace{}%
\AgdaOperator{\AgdaInductiveConstructor{,}}\AgdaSpace{}%
\AgdaBound{a}\AgdaSymbol{)}\AgdaSpace{}%
\AgdaOperator{\AgdaFunction{∈}}\AgdaSpace{}%
\AgdaBound{S}\AgdaSpace{}%
\AgdaSymbol{→}\AgdaSpace{}%
\AgdaBound{a}\AgdaSpace{}%
\AgdaOperator{\AgdaFunction{∈}}\AgdaSpace{}%
\AgdaBound{w}\AgdaSymbol{)}\AgdaSpace{}%
\AgdaOperator{\AgdaFunction{×}}\<%
\\
\>[.][@{}l@{}]\<[30I]%
\>[11]\AgdaSymbol{(∀}\AgdaSpace{}%
\AgdaBound{a}\AgdaSpace{}%
\AgdaSymbol{→}\AgdaSpace{}%
\AgdaSymbol{(}\AgdaInductiveConstructor{-}\AgdaSpace{}%
\AgdaOperator{\AgdaInductiveConstructor{,}}\AgdaSpace{}%
\AgdaBound{a}\AgdaSymbol{)}\AgdaSpace{}%
\AgdaOperator{\AgdaFunction{∈}}\AgdaSpace{}%
\AgdaBound{S}\AgdaSpace{}%
\AgdaSymbol{→}\AgdaSpace{}%
\AgdaBound{a}\AgdaSpace{}%
\AgdaOperator{\AgdaFunction{∉}}\AgdaSpace{}%
\AgdaBound{w}\AgdaSymbol{)}\<%
\end{code}

The rules of our calculus then say that it is safe to \agdaData{halt} a plan if our current world satisfies the goal state, and it is safe to sequence (\agdaData{seq}) another action to a plan if the current world satisfies the action's precondition. 
As expected, the rules of inference are then defined as an inductive
relation $\Gamma \vdash$ \agdaData{plan} $\ : \ $ \agdaData{initialWorld} $\strans$ \agdaData{goalState}: 


\begin{code}
\>[0]\AgdaKeyword{data}\AgdaSpace{}%
\AgdaOperator{\AgdaDatatype{\AgdaUnderscore{}⊢\AgdaUnderscore{}∶\AgdaUnderscore{}↝\AgdaUnderscore{}}}\AgdaSpace{}%
\AgdaSymbol{:}\AgdaSpace{}%
\AgdaFunction{Context}\AgdaSpace{}%
\AgdaSymbol{→}\AgdaSpace{}%
\AgdaDatatype{Plan}\AgdaSpace{}%
\AgdaSymbol{→}\AgdaSpace{}%
\AgdaFunction{World}
\AgdaSymbol{→}
\AgdaFunction{GoalState}\AgdaSpace{}%
\AgdaSymbol{→}\AgdaSpace{}%
\AgdaPrimitiveType{Set}\<%
\\
\>[0][@{}l@{\AgdaIndent{0}}]%
\>[2]\AgdaKeyword{where}\<%
\\
\>[2]\AgdaInductiveConstructor{halt}\AgdaSpace{}%
\AgdaSymbol{:}\AgdaSpace{}%
\AgdaSymbol{∀}\AgdaSpace{}\AgdaSymbol{\{}\AgdaBound{Γ}%
\>[74I]\AgdaBound{world}\AgdaSpace{}%
\AgdaBound{goal}\AgdaSymbol{\}}\AgdaSpace{}\<%
\\
\>[4]
\AgdaSymbol{→}\AgdaSpace{}%
\AgdaBound{world}\AgdaSpace{}%
\AgdaOperator{\AgdaFunction{∈⟨}}\AgdaSpace{}%
\AgdaBound{goal}\AgdaSpace{}%
\AgdaOperator{\AgdaFunction{⟩}}\<%
\\
\>[4]\AgdaSymbol{→}\AgdaSpace{}%
\AgdaBound{Γ}\AgdaSpace{}%
\AgdaOperator{\AgdaDatatype{⊢}}\AgdaSpace{}%
\AgdaInductiveConstructor{halt}\AgdaSpace{}%
\AgdaOperator{\AgdaDatatype{∶}}\AgdaSpace{}%
\AgdaBound{world}\AgdaSpace{}%
\AgdaOperator{\AgdaDatatype{↝}}\AgdaSpace{}%
\AgdaBound{goal}\<%
\\
\>[2]\AgdaInductiveConstructor{seq}\AgdaSpace{}%
\AgdaSymbol{:}\AgdaSpace{}%
\AgdaSymbol{∀}\AgdaSpace{}\AgdaSymbol{\{}
\AgdaBound{α}\AgdaSpace{}%
\AgdaBound{world}\AgdaSpace{}%
\AgdaBound{goal}\AgdaSpace{}%
\AgdaBound{Γ}\AgdaSpace{}%
\AgdaBound{f}\AgdaSymbol{\}}\<%
\\
\>[2][@{}l@{\AgdaIndent{0}}]%
\>[4]\AgdaSymbol{→}%
\>[7]\AgdaBound{world}%
\>[14]\AgdaOperator{\AgdaFunction{∈⟨}}\AgdaSpace{}%
\AgdaField{preconditions}\AgdaSpace{}%
\AgdaSymbol{(}\AgdaBound{Γ}\AgdaSpace{}%
\AgdaBound{α}\AgdaSymbol{)}\AgdaSpace{}%
\AgdaOperator{\AgdaFunction{⟩}}\<%
\\
\>[4]\AgdaSymbol{→}\AgdaSpace{}%
\AgdaBound{Γ}\AgdaSpace{}%
\AgdaOperator{\AgdaDatatype{⊢}}\AgdaSpace{}%
\AgdaBound{f}\AgdaSpace{}%
\AgdaOperator{\AgdaDatatype{∶}}\AgdaSpace{}%
\AgdaFunction{updateWorld}\AgdaSpace{}%
\AgdaSymbol{(}\AgdaField{effects}\AgdaSpace{}%
\AgdaSymbol{(}\AgdaBound{Γ}\AgdaSpace{}%
\AgdaBound{α}\AgdaSymbol{))}\AgdaSpace{}%
\AgdaBound{world}\AgdaSpace{}%
\AgdaOperator{\AgdaDatatype{↝}}\AgdaSpace{}%
\AgdaBound{goal}\<%
\\
\>[4]\AgdaSymbol{→}\AgdaSpace{}%
\AgdaBound{Γ}\AgdaSpace{}%
\AgdaOperator{\AgdaDatatype{⊢}}\AgdaSpace{}%
\AgdaSymbol{(}\AgdaBound{α}\AgdaSpace{}%
\AgdaOperator{\AgdaInductiveConstructor{∷}}\AgdaSpace{}%
\AgdaBound{f}\AgdaSymbol{)}\AgdaSpace{}%
\AgdaOperator{\AgdaDatatype{∶}}\AgdaSpace{}%
\AgdaBound{world}\AgdaSpace{}%
\AgdaOperator{\AgdaDatatype{↝}}\AgdaSpace{}%
\AgdaBound{goal}\<%
\end{code}

The actual sequencing is performed by the \agdaFunction{updateWorld} function, that applies the effects of an action by adding all the positively mapped predicates to the world and removing all the negatively mapped predicates. It should be noted that we do not need check the preconditions of an action in the \agdaFunction{updateWorld} function as we only apply actions on a world after ensuring that the world satisfies the preconditions of the action in the typing relation.

\begin{code}
\>[0]\AgdaFunction{updateWorld}\AgdaSpace{}%
\AgdaSymbol{:}\AgdaSpace{}%
\AgdaFunction{Effects}\AgdaSpace{}%
\AgdaSymbol{→}\AgdaSpace{}%
\AgdaFunction{World}\AgdaSpace{}%
\AgdaSymbol{→}\AgdaSpace{}%
\AgdaFunction{World}\<%
\\
\>[0]\AgdaFunction{updateWorld}\AgdaSpace{}%
\AgdaInductiveConstructor{[]}\AgdaSpace{}%
\>[20]\AgdaBound{w}\AgdaSpace{}%
\AgdaSymbol{=}\AgdaSpace{}%
\AgdaBound{w}\<%
\\
\>[0]\AgdaFunction{updateWorld}\AgdaSpace{}%
\AgdaSymbol{((}\AgdaInductiveConstructor{+}\AgdaSpace{}%
\AgdaOperator{\AgdaInductiveConstructor{,}}\AgdaSpace{}%
\AgdaBound{p}\AgdaSymbol{)}\AgdaSpace{}%
\AgdaOperator{\AgdaInductiveConstructor{∷}}\AgdaSpace{}%
\AgdaBound{S}\AgdaSymbol{)}\AgdaSpace{}%
\>[20]\AgdaBound{w}\AgdaSpace{}%
\AgdaSymbol{=}\AgdaSpace{}%
\AgdaBound{p}\AgdaSpace{}%
\AgdaOperator{\AgdaInductiveConstructor{∷}}\AgdaSpace{}%
\AgdaFunction{updateWorld}\AgdaSpace{}%
\AgdaBound{S}\AgdaSpace{}%
\AgdaBound{w}\<%
\\
\>[0]\AgdaFunction{updateWorld}\AgdaSpace{}%
\AgdaSymbol{((}\AgdaInductiveConstructor{-}\AgdaSpace{}%
\AgdaOperator{\AgdaInductiveConstructor{,}}\AgdaSpace{}%
\AgdaBound{p}\AgdaSymbol{)}\AgdaSpace{}%
\AgdaOperator{\AgdaInductiveConstructor{∷}}\AgdaSpace{}%
\AgdaBound{S}\AgdaSymbol{)}\AgdaSpace{}%
\>[20]\AgdaBound{w}\AgdaSpace{}%
\AgdaSymbol{=}\AgdaSpace{}%
\AgdaFunction{remove}\AgdaSpace{}%
\AgdaBound{p}\AgdaSpace{}%
\AgdaSymbol{(}\AgdaFunction{updateWorld}\AgdaSpace{}%
\AgdaBound{S}\AgdaSpace{}%
\AgdaBound{w}\AgdaSymbol{)}\<%
\end{code}

In \cite{SchwaabKHFPWH19}, these rules were proven sound and complete relative to the possible world semantics of PDDL. Compared to \cite{SchwaabKHFPWH19} we have simplified the rules so that we use a \agdaData{World} rather than a \agdaData{State} to represent the initial state of the problem.
Technically speaking, this is all we need to validate the intrinsic properties of a PDDL plan.
Since the rules are so simple, it is possible to generate Agda proofs automatically from PDDL plans, which we implemented as a function \agdaFunction{solver} in~\cite{HillKP20}.
Since the rules are defined generically, a user who works on a specific plan validation does not have to do anything, except for supplying a generic validation command (in which they insert the given plan as well as initial world and goal state):

\begin{code}
\>[0]\AgdaFunction{derivation}\AgdaSpace{}%
\AgdaSymbol{:}\AgdaSpace{}%
\AgdaFunction{Γ}\AgdaSpace{}%
\AgdaOperator{\AgdaDatatype{⊢}}\AgdaSpace{}%
\AgdaFunction{plan}\AgdaSpace{}%
\AgdaOperator{\AgdaDatatype{∶}}\AgdaSpace{}%
\AgdaFunction{initialWorld}\AgdaSpace{}%
\AgdaOperator{\AgdaDatatype{↝}}\AgdaSpace{}%
\AgdaFunction{goalState}\<%
\\
\>[0]\AgdaFunction{derivation}\AgdaSpace{}%
\AgdaSymbol{=}\AgdaSpace{}%
\AgdaFunction{from-just}\AgdaSpace{}%
\AgdaSymbol{(}\AgdaFunction{solver}\AgdaSpace{}%
\AgdaFunction{Γ}\AgdaSpace{}%
\AgdaFunction{plan}\AgdaSpace{}%
\AgdaFunction{initialWorld}\AgdaSpace{}%
\AgdaFunction{goalState}\AgdaSymbol{)}\<%
\end{code}

The initial motivation behind this work was in giving \emph{Curry-Howard}, or \emph{computational} interpretation to AI plans, with a view of opening the way to a certified code extraction. With that in mind, the inference rules defining
\begin{equation*}
 \Gamma \vdash \text{\agdaData{plan}} \ : \  \text{\agdaData{initialWorld}} \strans \text{\agdaData{goalState}}
\end{equation*}
 model plans as functions that inhabit the type
\begin{equation*} 
\text{\agdaData{initialWorld}} \strans \text{\agdaData{goalState}}
\end{equation*}
In Section~\ref{sec:future} we also discuss how this approach differs from modelling PPDP plans within linear logic.  

\subsection{Plan Execution: Action Handlers}\label{sec:handlers}

In \cite{SchwaabKHFPWH19} we introduced the notion of a \emph{canonical action handler}, that can take a plan validated as in previous section, and turn it into an executable function over the \emph{possible worlds}, as defined in PDDL semantics. 

In order to discuss our approach of verifying extrinsic properties, only the notion of the possible \agdaData{World} is relevant. We refer interested readers to ~\cite{SchwaabKHFPWH19} for a complete definition of the possible world semantics.

A handler executes \agdaData{Action}s on \AgdaFunction{World}s:
\begin{code}
\>[0]\AgdaFunction{ActionHandler}\AgdaSpace{}%
\AgdaSymbol{:}\AgdaSpace{}%
\AgdaPrimitiveType{Set}\<%
\\
\>[0]\AgdaFunction{ActionHandler}\AgdaSpace{}%
\AgdaSymbol{=}\AgdaSpace{}%
\AgdaField{Action}\AgdaSpace{}%
\AgdaSymbol{$\rightarrow$}\AgdaSpace{}%
\AgdaFunction{World}\AgdaSpace{}%
\AgdaSymbol{$\rightarrow$}\AgdaSpace{}%
\AgdaFunction{World}\<%
\end{code}

We now define a \emph{canonical handler} which applies the effects of an action according to the context by using the \agdaFunction{updateWorld} function.

\begin{code}
[\AgdaEmptyExtraSkip]%
\>[0]\AgdaFunction{canonical-σ}\AgdaSpace{}%
\AgdaSymbol{:}\AgdaSpace{}%
\AgdaFunction{Context}\AgdaSpace{}%
\AgdaSymbol{→}\AgdaSpace{}%
\AgdaFunction{ActionHandler}\<%
\\
\>[0]\AgdaFunction{canonical-σ}\AgdaSpace{}%
\AgdaBound{Γ}\AgdaSpace{}%
\AgdaBound{α}\AgdaSpace{}%
\AgdaSymbol{=}\AgdaSpace{}%
\AgdaFunction{updateWorld}\AgdaSpace{}%
\AgdaSymbol{(}\AgdaFunction{effects}\AgdaSpace{}%
\AgdaSymbol{(}\AgdaBound{Γ}\AgdaSpace{}%
\AgdaBound{α}\AgdaSymbol{))}\<%
\end{code}

To be able to evaluate an entire plan we define an \AgdaFunction{execute} function that takes in a plan, action handler and initial world as its arguments and recursively applies all actions in the plan using the given action handler to the world until the end of the plan.
\begin{code}
\>[0]\AgdaFunction{execute}\AgdaSpace{}%
\AgdaSymbol{:}\AgdaSpace{}%
\AgdaFunction{Plan}\AgdaSpace{}%
\AgdaSymbol{→}\AgdaSpace{}%
\AgdaFunction{ActionHandler}\AgdaSpace{}%
\AgdaSymbol{→}\AgdaSpace{}%
\AgdaFunction{World}\AgdaSpace{}%
\AgdaSymbol{→}\AgdaSpace{}%
\AgdaFunction{World}\<%
\\
\>[0]\AgdaFunction{execute}\AgdaSpace{}%
\AgdaInductiveConstructor{halt}\AgdaSpace{}%
\>[20]\AgdaBound{σ}\AgdaSpace{}%
\AgdaBound{w}\AgdaSpace{}%
\AgdaSymbol{=}\AgdaSpace{}%
\AgdaBound{w}\<%
\\
\>[0]\AgdaFunction{execute}\AgdaSpace{}%
\AgdaSymbol{(}\AgdaBound{α}\AgdaSpace{}%
\AgdaOperator{\AgdaInductiveConstructor{∷}}\AgdaSpace{}%
\AgdaBound{f}\AgdaSymbol{)}\AgdaSpace{}%
\>[20]\AgdaBound{σ}\AgdaSpace{}%
\AgdaBound{w}\AgdaSpace{}%
\AgdaSymbol{=}\AgdaSpace{}%
\AgdaFunction{execute}\AgdaSpace{}%
\AgdaBound{f}\AgdaSpace{}%
\AgdaBound{σ}\AgdaSpace{}%
\AgdaSymbol{(}\AgdaBound{σ}\AgdaSpace{}%
\AgdaBound{α}\AgdaSpace{}%
\AgdaBound{w}\AgdaSymbol{)}\<%
\end{code}
Note that in this case \AgdaFunction{execute} could simply be defined as a fold over the list of actions. We have left it in this explicit form, as in the next section we will alter the definition of \AgdaFunction{ActionHandler} to use dependent types in order to encode rich extrinsic properties, which means that expressing this as a fold is no longer possible.

We can now evaluate the taxi example by applying the \AgdaFunction{execute} function to the canonical handler and initial world. 
\begin{code}
\>[0]\AgdaFunction{evaluationResult}\AgdaSpace{}%
\AgdaSymbol{:}\AgdaSpace{}%
\AgdaFunction{World}\<%
\\
\>[0]\AgdaFunction{evaluationResult}\AgdaSpace{}%
\AgdaSymbol{=}\AgdaSpace{}%
\AgdaFunction{execute}\AgdaSpace{}%
\AgdaFunction{plan}\AgdaSpace{}%
\AgdaSymbol{(}\AgdaFunction{canonical-σ}\AgdaSpace{}%
\AgdaFunction{Γ}\AgdaSymbol{)}\AgdaSpace{}%
\AgdaFunction{initialWorld}\<%
\end{code}
As we execute the already validated plan on the \agdaData{initialWorld}, we expect to see, as an output, a world that corresponds to the goal state \agdaData{goalState}.
In fact, we get:
\begin{code}
\>[0]\AgdaComment{Output: }\<
\\
\>[0]\AgdaComment{   taxiIn taxi3 location3 ∷ }\<
\\
\>[0]\AgdaComment{ personIn person1 location3 ∷ }\<
\\
\>[0]\AgdaComment{   personIn person3 location1 ∷ }\<
\\
\>[0]\AgdaComment{   taxiIn taxi1 location2 ∷ }\<
\\
\>[0]\AgdaComment{   taxiIn taxi2 location2 ∷ }\<
\\
\>[0]\AgdaComment{   personIn person2 location2 ∷ [] }\<
\end{code}
That is, the world that the function \agdaFunction{evaluationResult} returns is larger than the world the \agdaData{goalState} directly entails, but this is expected, as long as the information contained in  \agdaData{goalState} is preserved.

Note that generally, given a state, there may be many worlds that satisfy it. For example, the following world also satisfies our \agdaData{goalState}.
\begin{code}
\>[0]\AgdaComment{   taxiIn taxi3 location3 ∷ }\<
\\
\>[0]\AgdaComment{ personIn person1 location3 ∷ }\<
\\
\>[0]\AgdaComment{   personIn person3 location1 ∷ []}\<
\end{code}
  


The central \emph{soundness} result of~\cite{SchwaabKHFPWH19} states that, whenever we can prove that
$\Gamma \vdash$ \agdaData{plan} \ : \ \agdaData{initialWorld} $\strans$ \agdaData{goalState}
then executing the \agdaData{plan} on \agdaData{initialWorld} will result in a world $w'$ that satisfies conditions in the \agdaData{goalState}.

  We rely on our proof of plan soundness to establish that the plan, and therefore each action in it, is valid. As a consequence we do not require action handlers to check the validity of each action with respect to the current world before applying it. This allows us to simplify the definition of the enriched handlers, described in the next section. 

This finishes the recap of canonical handlers from \cite{SchwaabKHFPWH19}, we are now ready to tackle extrinsic properties in the next section.

\section{Verifying extrinsic properties}\label{sec:policies}

So far we have introduced two out of three components of our proposed framework:

\begin{itemize}
\item[(I)] \textbf{Plan generation via a PDDL planner} which takes a PDDL domain and problem definition as in Figures~\ref{fig:pddl-taxi-domain}~\&~\ref{fig:pddl-taxi-problem}, and performs an automated search for plan that takes the system from the initial to the goal state.
  \item[(II)] \textbf{Validation of the resulting plan via Agda} which compiles the planning domain, planning problem and plan received from the planner into a compact DSL. The plan is then validated relative to the formalised operational and possible world semantics of PDDL. We have shown in ~\cite{HillKP20} that this validation stage is capable of eliminating state inconsistencies that are otherwise admitted by the STRIPS planner.
\end{itemize}
We are now ready to introduce a third, and perhaps the most intriguing component of the framework:
\begin{itemize}
\item[(III)] \textbf{Dependently-typed verification of extrinsic properties of the execution of the plan via Agda}, in which we formally verify that during execution of the plan the controller will never execute an action that violates extrinsic properties. As discussed previously, extrinsic properties are those which are either undesirable or impossible to encode in PDDL at planning time.
\end{itemize}
To achieve this, we augment the \AgdaFunction{ActionHandler} type with the desired property and then ensure that the \AgdaFunction{execute} function has the correct type. Note that this third stage lacks the generality of the stage (II), as these higher level properties are necessarily specific to the particular domain being modelled. Nonetheless we argue it is a powerful and flexible technique for verifying properties that cannot be checked at planning time. A notable advantage of our approach is that we can verify a property holds without altering the semantics of PDDL specification or the shape of the plans produced by the planner.

As discussed in the introduction, the verified controller could be used directly during execution. Alternatively, it could be run in a simulated environment prior to the execution of the plan by a non-verified controller. If the plan satisfies the extrinsic properties during simulation, then the environment can be monitored during execution and as long as it doesn't deviate from the simulation then the actual execution of the plan will not violate the extrinsic properties either.

%

\subsection{Example 1: Fuel Consumption}

One very simple property that we might be interested in verifying is that the agent never runs out of fuel while executing a plan. Although fuel is often used in an abstract sense in functional programming to limit the number iterations a function may perform before termination, in planning fuel often has a very real interpretation as it represents a resource (e.g. electrical energy) that an agent uses to perform actions. Typically, before an agent runs out of fuel it must return to its base and recharge. 

In many domains, fuel levels cannot be taken into account by the planner in stage (I) because it is unknown what the exact fuel level will be at a given point in the plan. For example while the plan of driving from \AgdaInductiveConstructor{location1} to \AgdaInductiveConstructor{location2} and then from \AgdaInductiveConstructor{location2} to \AgdaInductiveConstructor{location3} may be valid at planning time, it can subsequently be invalidated by unexpectedly high fuel consumption during the first leg of the trip (e.g. due to a diversion caused by road-works) that leaves the taxi unable to complete the second leg. 

We will now show how such a constraint can be incorporated into our framework. For simplicity and pedagogical purposes we will assume that all taxis share a single fuel source and that applying any action uses $1$ unit of fuel. To add this property to an action handler we first model the property using types in Agda and then add that property to the type level of the action handler.
To allow us to reason about natural numbers at the type level we create a \agdaData{Fuel} data type that is indexed by a natural number:

\begin{code}
\>[0]\AgdaKeyword{data}\AgdaSpace{}%
\AgdaDatatype{Fuel}\AgdaSpace{}%
\AgdaSymbol{:}\AgdaSpace{}%
\AgdaDatatype{Nat}\AgdaSpace{}%
\AgdaSymbol{→}\AgdaSpace{}%
\AgdaPrimitiveType{Set}\AgdaSpace{}%
\AgdaKeyword{where}\<%
\\
\>[0][@{}l@{\AgdaIndent{0}}]%
\>[2]\AgdaInductiveConstructor{fuel}\AgdaSpace{}%
\AgdaSymbol{:}\AgdaSpace{}%
\AgdaSymbol{(}\AgdaBound{n}\AgdaSpace{}%
\AgdaSymbol{:}\AgdaSpace{}%
\AgdaDatatype{Nat}\AgdaSymbol{)}\AgdaSpace{}%
\AgdaSymbol{→}\AgdaSpace{}%
\AgdaDatatype{Fuel}\AgdaSpace{}%
\AgdaBound{n}\<%
\end{code}
We can now enrich the definition of an action handler, by encoding the fact that applying an action reduces the fuel level from \agdaFunction{suc} $n$ to $n$ where \agdaFunction{suc} $n$ is the successor of the natural number $n$. 
\begin{code}
\>[0]\AgdaFunction{FuelAwareActionHandler}\AgdaSpace{}%
\AgdaSymbol{=}\AgdaSpace{}%
\AgdaSymbol{∀}\AgdaSpace{}%
\AgdaSymbol{\{}\AgdaBound{n}\AgdaSymbol{\}}\AgdaSpace{}%
\AgdaSymbol{→}\AgdaSpace{}%
\AgdaDatatype{Action}\<%
\\
\>[0][@{}l@{\AgdaIndent{0}}]%
\>[22]\AgdaSymbol{→}\AgdaSpace{}%
\AgdaFunction{World}\AgdaSpace{}%
\AgdaOperator{\AgdaFunction{×}}\AgdaSpace{}%
\AgdaDatatype{Fuel}\AgdaSpace{}%
\AgdaSymbol{(}\AgdaInductiveConstructor{suc}\AgdaSpace{}%
\AgdaBound{n}\AgdaSymbol{)}\<%
\\
\>[22]\AgdaSymbol{→}\AgdaSpace{}%
\AgdaFunction{World}\AgdaSpace{}%
\AgdaOperator{\AgdaFunction{×}}\AgdaSpace{}%
\AgdaDatatype{Fuel}\AgdaSpace{}%
\AgdaBound{n}\<%
\end{code}
This encodes at the type level that the agent cannot begin to execute an action without having sufficient energy and that each action uses one unit of fuel. 

Now we can define an \emph{enriched handler}, by changing the canonical handler's return type to  \agdaFunction{FuelAwareActionHandler}:
\begin{code}
\>[0]\AgdaFunction{enriched-σ}\AgdaSpace{}%
\AgdaSymbol{:}\AgdaSpace{}%
\AgdaFunction{Context}\AgdaSpace{}%
\AgdaSymbol{→}\AgdaSpace{}%
\AgdaFunction{FuelAwareActionHandler}\<%
\\
\>[0]\AgdaFunction{enriched-σ}\AgdaSpace{}%
\AgdaBound{Γ}\AgdaSpace{}%
\AgdaBound{α}\AgdaSpace{}%
\AgdaSymbol{=}\AgdaSpace{}%
\AgdaFunction{updateWorld'}\AgdaSpace{}%
\AgdaSymbol{(}\AgdaFunction{effects}\AgdaSpace{}%
\AgdaSymbol{(}\AgdaBound{Γ}\AgdaSpace{}%
\AgdaBound{α}\AgdaSpace{}%
\AgdaSymbol{))}\<%
\end{code}
The auxiliary function \AgdaFunction{updateWorld'} above is an enriched version of \AgdaFunction{updateWorld} we used in the Section~\ref{sec:handlers}. It takes care of checking the additional \agdaFunction{fuel} constraint on the type of the action handler is satisfied during the execution:
\begin{code}
\>[0]\AgdaFunction{updateWorld'}\AgdaSpace{}%
\AgdaSymbol{:}\AgdaSpace{}%
\AgdaFunction{Effect}\AgdaSpace{}%
\AgdaSymbol{→}%
\>[24]\AgdaFunction{World}\AgdaSpace{}%
\AgdaOperator{\AgdaFunction{×}}\AgdaSpace{}%
\AgdaDatatype{Fuel}\AgdaSpace{}%
\AgdaSymbol{(}\AgdaInductiveConstructor{suc}\AgdaSpace{}%
\AgdaGeneralizable{n}\AgdaSymbol{)}\<%
\\
\>[0][@{}l@{\AgdaIndent{0}}]%
\>[12]\AgdaSymbol{→}\AgdaSpace{}%
\AgdaFunction{World}\AgdaSpace{}%
\AgdaOperator{\AgdaFunction{×}}\AgdaSpace{}%
\AgdaDatatype{Fuel}\AgdaSpace{}%
\AgdaGeneralizable{n}\<%
\\
\>[0]\AgdaFunction{updateWorld'}\AgdaSpace{}%
\AgdaInductiveConstructor{s}\AgdaSpace{}%
\AgdaSymbol{(}\AgdaBound{w}\AgdaSpace{}%
\AgdaOperator{\AgdaInductiveConstructor{,}}\AgdaSpace{}%
\AgdaInductiveConstructor{fuel}\AgdaSpace{}%
\AgdaSymbol{(}\AgdaInductiveConstructor{suc}\AgdaSpace{}%
\AgdaBound{n}\AgdaSymbol{))}
\AgdaSymbol{=}\AgdaSpace{}%
\AgdaFunction{updateWorld}\AgdaSpace{}%
\AgdaBound{S}\AgdaSpace{}%
\AgdaBound{w}\AgdaSpace{}%
\AgdaOperator{\AgdaInductiveConstructor{,}}\AgdaSpace{}%
\AgdaInductiveConstructor{fuel}\AgdaSpace{}%
\AgdaBound{n}\<%
\end{code}
One advantage of defining at the type level that the fuel goes from \agdaFunction{suc} $n$ to $n$ is
that we are forced to supply an energy of exactly $n$ in the return type of this function.

To execute plans with the \agdaFunction{FuelAwareActionHandler} and check the constraints during execution,
we need to further enrich the evaluation function.
 The evaluation function must check the \agdaFunction{fuel} level and if it is \agdaFunction{suc} $n$ we \emph{handle} the action and if it is \agdaFunction{zero} whilst there are still actions to apply then the plan fails in which case we return an error message with the failure. One simple way to implement this is to introduce a disjunction \textcolor{AgdaDatatype}{$\uplus$} in the return type where the function can either return a world or an error if the execution fails. To do this we define a \agdaData{OutOfFuelError} data type that is constructed by passing in the current world and the failed action.
\begin{code}
\>[0]\AgdaKeyword{data}\AgdaSpace{}%
\AgdaDatatype{OutOfFuelError}\AgdaSpace{}%
\AgdaSymbol{:}\AgdaSpace{}%
\AgdaPrimitiveType{Set}\AgdaSpace{}%
\AgdaKeyword{where}\<%
\\
\>[0][@{}l@{\AgdaIndent{0}}]%
\>[2]\AgdaInductiveConstructor{error}\AgdaSpace{}%
\AgdaSymbol{:}\AgdaSpace{}%
\AgdaDatatype{Action}\AgdaSpace{}%
\AgdaSymbol{→}\AgdaSpace{}%
\AgdaFunction{World}\AgdaSpace{}%
\AgdaSymbol{→}\AgdaSpace{}%
\AgdaDatatype{OutOfFuelError}\<%
\\
\\
\>[0]\AgdaFunction{executeWithFuel}\AgdaSpace{}%
\AgdaSymbol{:}\AgdaSpace{}%
\AgdaDatatype{Plan}\AgdaSpace{}%
\AgdaSymbol{→}\AgdaSpace{}%
\AgdaFunction{FuelAwareActionHandler}\<%
\\
\>[0][@{}l@{\AgdaIndent{0}}]%
\>[13]\AgdaSymbol{→}\AgdaSpace{}%
\AgdaFunction{World}\AgdaSpace{}%
\AgdaOperator{\AgdaFunction{×}}\AgdaSpace{}%
\AgdaDatatype{Fuel}\AgdaSpace{}%
\AgdaGeneralizable{n}\<%
\\
\>[13]\AgdaSymbol{→}\AgdaSpace{}%
\AgdaFunction{World}\AgdaSpace{}%
\AgdaOperator{\AgdaDatatype{⊎}}\AgdaSpace{}%
\AgdaDatatype{OutOfFuelError}\<%
\\
\>[0]\AgdaCatchallClause{\AgdaFunction{executeWithFuel}}\AgdaSpace{}%
\AgdaCatchallClause{\AgdaInductiveConstructor{halt}}\AgdaSpace{}%
\>[60]\AgdaCatchallClause{\AgdaBound{σ}}\AgdaSpace{}%
\AgdaCatchallClause{\AgdaSymbol{(}}\AgdaCatchallClause{\AgdaBound{w}}\AgdaSpace{}%
\AgdaCatchallClause{\AgdaOperator{\AgdaInductiveConstructor{,}}}\AgdaSpace{}%
\AgdaCatchallClause{\AgdaBound{\_}}\AgdaCatchallClause{\AgdaSymbol{)}}\AgdaSpace{}%
\AgdaSymbol{=}\AgdaSpace{}%
\AgdaInductiveConstructor{inj$_1$}\AgdaSpace{}%
\AgdaBound{w}\<%
\\
\>[0]\AgdaFunction{executeWithFuel}\AgdaSpace{}%
\AgdaSymbol{(}\AgdaBound{α}\AgdaSpace{}%
\AgdaOperator{\AgdaInductiveConstructor{∷}}\AgdaSpace{}%
\AgdaBound{f}\AgdaSymbol{)}\AgdaSpace{}%
\>[60]\AgdaBound{σ}\AgdaSpace{}%
\AgdaSymbol{(}\AgdaBound{w}\AgdaSpace{}%
\AgdaOperator{\AgdaInductiveConstructor{,}}\AgdaSpace{}%
\AgdaInductiveConstructor{fuel}\AgdaSpace{}%
\AgdaInductiveConstructor{0}\AgdaSymbol{)}\AgdaSpace{}%
\AgdaSymbol{=}\AgdaSpace{}%
\AgdaInductiveConstructor{inj$_2$}\AgdaSpace{}%
\AgdaSymbol{(}\AgdaInductiveConstructor{error}\AgdaSpace{}%
\AgdaBound{α}\AgdaSpace{}%
\AgdaBound{w}\AgdaSymbol{)}\<%
\\
\>[0]\AgdaFunction{executeWithFuel}\AgdaSpace{}%
\AgdaSymbol{(}\AgdaBound{α}\AgdaSpace{}%
\AgdaOperator{\AgdaInductiveConstructor{∷}}\AgdaSpace{}%
\AgdaBound{f}\AgdaSymbol{)}\AgdaSpace{}%
\>[60]\AgdaBound{σ}\AgdaSpace{}%
\AgdaSymbol{(}\AgdaBound{w}\AgdaSpace{}%
\AgdaOperator{\AgdaInductiveConstructor{,}}\AgdaSpace{}%
\AgdaInductiveConstructor{fuel}\AgdaSpace{}%
\AgdaSymbol{(}\AgdaInductiveConstructor{suc}\AgdaSpace{}%
\AgdaBound{n}\AgdaSymbol{))}\AgdaSpace{}%
\AgdaSymbol{=}\<%
\\
\>[0][@{}l@{\AgdaIndent{0}}]%
\>[12]\AgdaFunction{executeWithFuel}\AgdaSpace{}%
\AgdaBound{f}\AgdaSpace{}%
\AgdaBound{σ}\AgdaSpace{}%
\AgdaSymbol{(}\AgdaBound{σ}\AgdaSpace{}%
\AgdaBound{α}\AgdaSpace{}%
\AgdaSymbol{(}\AgdaBound{w}\AgdaSpace{}%
\AgdaOperator{\AgdaInductiveConstructor{,}}\AgdaSpace{}%
\AgdaInductiveConstructor{fuel}\AgdaSpace{}%
\AgdaSymbol{(}\AgdaInductiveConstructor{suc}\AgdaSpace{}%
\AgdaBound{n}\AgdaSymbol{)))}\<%
\end{code}
We can now execute the same \agdaData{plan} that we validated in the previous section,
only this time we have the \agdaData{enriched} (rather than \agdaData{canonical}) handler and evaluation function:

\begin{code}
\>[0]\AgdaFunction{evaluationResult}\AgdaSpace{}%
\AgdaSymbol{:}\AgdaSpace{}%
\AgdaFunction{World}\AgdaSpace{}%
\AgdaOperator{\AgdaDatatype{⊎}}\AgdaSpace{}%
\AgdaDatatype{OutOfFuelError}\<%
\\
\>[0]\AgdaFunction{evaluationResult}\AgdaSpace{}%
\AgdaSymbol{=}\AgdaSpace{}%
\AgdaFunction{executeWithFuel}%
\>[190I]\AgdaFunction{plan}\AgdaSpace{}%
\AgdaSymbol{(}\AgdaFunction{enriched-σ}\AgdaSpace{}%
\AgdaFunction{Γ}\AgdaSymbol{)}\<%
\\
\>[.][@{}l@{}]\<[190I]%
\>[30]\AgdaSymbol{(}\AgdaFunction{initialWorld}\AgdaSpace{}%
\AgdaOperator{\AgdaInductiveConstructor{,}}\AgdaSpace{}%
\AgdaSymbol{(}\AgdaInductiveConstructor{fuel}\AgdaSpace{}%
\AgdaNumber{3}\AgdaSymbol{))}\<%
\end{code}

This section used a simple fuel consumption example to explain the general approach of reasoning about meta-properties of already validated plans and demonstrated how enriched handlers allow us to introduce arbitrary additional constraints at execution time without interfering with either the native (sound) semantics of PDDL, or the shape of the native plans produced by STRIPS. In a realistic system, fuel levels might be better implemented as a monadic evaluation function that performs real-time measurement of the current fuel level.

\subsection{Example 2: Fairness}
\label{sec:fairness-example}

We will now look at a more complex constraint, in particular that the assignment of taxi drivers to trips exhibits no significant gender bias. Unlike the fuel consumption example, the gender information could be made available to the planner at Stage (I). However it is infeasible and undesirable to do so for the following two reasons. Firstly, any non-trivial fairness  property is unlikely to be expressible in standard PDDL syntax. Secondly and perhaps more subtly, statistically speaking we would expect there to be no gender bias in the output of the planner in the first place. The time complexity of planning algorithms are normally non-linear in the size of the domain description, so we why complicate the planning stage to enforce something that should be normally true most of the time? Verifying that the property holds only at execution time significantly reduces the cost.

To encode this property in Agda we first need to define a model of gender in Agda. 
\begin{code}
\>[0]\AgdaKeyword{data}\AgdaSpace{}%
\AgdaDatatype{Gender}\AgdaSpace{}%
\AgdaSymbol{:}\AgdaSpace{}%
\AgdaPrimitiveType{Set}\AgdaSpace{}%
\AgdaKeyword{where}\<%
\\
\>[0][@{}l@{\AgdaIndent{0}}]%
\>[2]\AgdaInductiveConstructor{male}\AgdaSpace{}%
\AgdaInductiveConstructor{female}\AgdaSpace{}%
\AgdaInductiveConstructor{other}\AgdaSpace{}%
\AgdaSymbol{:}\AgdaSpace{}%
\AgdaDatatype{Gender}\<%
\end{code}
We then define a \agdaFunction{TripCount} type which is used to store the number of trips each gender has taken so far.
\begin{code}
\>[0]\AgdaFunction{TripCount}\AgdaSpace{}%
\AgdaSymbol{:}\AgdaSpace{}%
\AgdaPrimitiveType{Set}\<%
\\
\>[0]\AgdaFunction{TripCount}\AgdaSpace{}%
\AgdaSymbol{=}\AgdaSpace{}%
\AgdaDatatype{Gender}\AgdaSpace{}%
\AgdaSymbol{→}\AgdaSpace{}%
\AgdaDatatype{ℕ}\<%
\end{code}
We will define the code associated with the enriched handler in a separate Agda module. The advantage of this choice is that we can pass in static functions representing data that we do not intend to change during the evaluation of a given domain. 
\begin{code}
\>[20]\AgdaBound{driverGender}\AgdaSpace{}%
\AgdaSymbol{:}\AgdaSpace{}%
\AgdaDatatype{Object}\AgdaSpace{}%
\AgdaInductiveConstructor{taxi}\AgdaSpace{}%
\AgdaSymbol{→}\AgdaSpace{}%
\AgdaDatatype{Gender}\<%
\\
\>[20]\AgdaBound{margin}\AgdaSpace{}%
\AgdaSymbol{:}\AgdaSpace{}%
\AgdaDatatype{Nat}
\end{code}
In particular we pass in two functions \agdaFunction{driverGender} and one natural number called \agdaFunction{margin}. The \agdaFunction{driverGender} function maps all taxi drivers to a \agdaData{Gender}. The \agdaFunction{margin} is used to allow for some leeway for statistical fluctuations when enforcing our fairness constraint.

We simply calculate the total trips taken by adding the trips taken for all genders.

\begin{code}
\>[0]\AgdaFunction{totalTripsTaken}\AgdaSpace{}%
\AgdaSymbol{:}\AgdaSpace{}%
\AgdaFunction{TripCount}\AgdaSpace{}%
\AgdaSymbol{→}\AgdaSpace{}%
\AgdaDatatype{ℕ}\<%
\\
\>[0]\AgdaFunction{totalTripsTaken}\AgdaSpace{}%
\AgdaBound{f}\AgdaSpace{}%
\AgdaSymbol{=}\AgdaSpace{}%
\AgdaOperator{\AgdaPrimitive{\AgdaUnderscore{}+\AgdaUnderscore{}}}\AgdaSpace{}%
\AgdaSymbol{(}\AgdaOperator{\AgdaPrimitive{\AgdaUnderscore{}+\AgdaUnderscore{}}}\AgdaSpace{}%
\AgdaSymbol{(}\AgdaBound{f}\AgdaSpace{}%
\AgdaInductiveConstructor{male}\AgdaSymbol{)}\AgdaSpace{}%
\AgdaSymbol{(}\AgdaBound{f}\AgdaSpace{}%
\AgdaInductiveConstructor{female}\AgdaSymbol{))}\AgdaSpace{}%
\AgdaSymbol{(}\AgdaBound{f}\AgdaSpace{}%
\AgdaInductiveConstructor{other}\AgdaSymbol{)}\<%
\end{code}

The percentage of trips assigned to a given gender is then calculated via the following function:
\begin{code}
\>[0]\AgdaFunction{calculateGenderAssignment}\AgdaSpace{}%
\AgdaSymbol{:}\AgdaSpace{}%
\AgdaDatatype{Gender}\AgdaSpace{}%
\AgdaSymbol{→}\AgdaSpace{}%
\AgdaFunction{TripCount}\AgdaSpace{}%
\AgdaSymbol{→}\AgdaSpace{}%
\AgdaDatatype{ℕ}\<%
\\
\>[0]\AgdaFunction{calculateGenderAssignment}\AgdaSpace{}%
\AgdaBound{g}\AgdaSpace{}%
\AgdaBound{tripCount}\AgdaSpace{}%
\AgdaSymbol{=}\<%
\\
\>[0][@{}l@{\AgdaIndent{0}}]%
\>[2]\AgdaSymbol{(}\AgdaBound{tripCount}\AgdaSpace{}%
\AgdaBound{g}\AgdaSpace{}%
\AgdaOperator{\AgdaPrimitive{*}}\AgdaSpace{}%
\AgdaNumber{100}\AgdaSymbol{)}\AgdaSpace{}%
\AgdaOperator{\AgdaFunction{/₀}}\AgdaSpace{}%
\AgdaFunction{totalTripsTaken}\AgdaSpace{}%
\AgdaBound{tripCount}\<%
\end{code}
To calculate a fair percentage of assignments we first need to calculate the number of drivers of each gender. Note that this uses a higher order function \AgdaFunction{filter} which, as discussed in Section~\ref{sec:PDDL-expressivity}, are not supported by the PDDL language. Neither are we aware of any alternative way of expressing this calculation in PDDL short of providing the taxis of each gender manually, an approach which scale extremely poorly as the domain grew in size.
\begin{code}
\>[0]\AgdaFunction{noGender}\AgdaSpace{}%
\AgdaSymbol{:}\AgdaSpace{}%
\AgdaDatatype{Gender}\AgdaSpace{}%
\AgdaSymbol{→}\AgdaSpace{}%
\AgdaDatatype{ℕ}\<%
\\
\>[0]\AgdaFunction{noGender}\AgdaSpace{}%
\AgdaBound{g}\AgdaSpace{}%
\AgdaSymbol{=}\<%
\\
\>[0][@{}l@{\AgdaIndent{0}}]%
\>[2]\AgdaFunction{length}\AgdaSpace{}%
\AgdaSymbol{(}\AgdaFunction{filter}\AgdaSpace{}%
\AgdaSymbol{(λ}\AgdaSpace{}%
\AgdaBound{t}\AgdaSpace{}%
\AgdaSymbol{→}\AgdaSpace{}%
\AgdaFunction{decGender}\AgdaSpace{}%
\AgdaBound{g}\AgdaSpace{}%
\AgdaSymbol{(}\AgdaBound{driverGender}\AgdaSpace{}%
\AgdaBound{t}\AgdaSymbol{))}\AgdaSpace{}%
\AgdaFunction{allTaxis}\AgdaSymbol{)}\<%
\end{code}
Using this we can then calculate the percentage of drivers of a given gender:
\begin{code}
\>[0]\AgdaFunction{percentage}\AgdaSpace{}%
\AgdaSymbol{:}\AgdaSpace{}%
\AgdaDatatype{Gender}\AgdaSpace{}%
\AgdaSymbol{→}\AgdaSpace{}%
\AgdaDatatype{ℕ}\<%
\\
\>[0]\AgdaFunction{percentage}\AgdaSpace{}%
\AgdaBound{g}\AgdaSpace{}%
\AgdaSymbol{=}\AgdaSpace{}%
\AgdaSymbol{(}\AgdaFunction{noGender}\AgdaSpace{}%
\AgdaBound{g}\AgdaSpace{}%
\AgdaOperator{\AgdaPrimitive{*}}\AgdaSpace{}%
\AgdaNumber{100}\AgdaSymbol{)}\AgdaSpace{}%
\AgdaOperator{\AgdaFunction{/₀}}\AgdaSpace{}%
\AgdaFunction{totalDrivers}\<%
\end{code}
The lowest acceptable threshold that is deemed to be fair, which is controlled by a \AgdaBound{margin} parameter, is then calculated as follows:
\begin{code}
\>[0]\AgdaFunction{calculateLowerbound}\AgdaSpace{}%
\AgdaSymbol{:}\AgdaSpace{}%
\AgdaDatatype{Gender}\AgdaSpace{}%
\AgdaSymbol{→}\AgdaSpace{}%
\AgdaDatatype{ℕ}\<%
\\
\>[0]\AgdaFunction{calculateLowerbound}\AgdaSpace{}%
\AgdaBound{g}\AgdaSpace{}%
\AgdaSymbol{=}\<%
\\
\>[0][@{}l@{\AgdaIndent{0}}]%
\>[3]\AgdaFunction{percentage}\AgdaSpace{}%
\AgdaBound{g}\AgdaSpace{}%
\AgdaOperator{\AgdaPrimitive{∸}}\AgdaSpace{}%
\AgdaSymbol{(}\AgdaFunction{percentage}\AgdaSpace{}%
\AgdaBound{g}\AgdaSpace{}%
\AgdaOperator{\AgdaFunction{/₀}}\AgdaSpace{}%
\AgdaBound{margin}\AgdaSymbol{)}\<%
\end{code}
We can now express the property that a trip count is unbiased for a particular gender as follows:
\begin{code}
\>[0]\AgdaFunction{IsFair}\AgdaSpace{}%
\AgdaSymbol{:}\AgdaSpace{}%
\AgdaDatatype{Gender}\AgdaSpace{}%
\AgdaSymbol{→}\AgdaSpace{}%
\AgdaFunction{TripCount}\AgdaSpace{}%
\AgdaSymbol{→}\AgdaSpace{}%
\AgdaPrimitiveType{Set}\<%
\\
\>[0]\AgdaFunction{IsFair}\AgdaSpace{}%
\AgdaBound{g}\AgdaSpace{}%
\AgdaBound{f}\AgdaSpace{}%
\AgdaSymbol{=}\<%
\\
\>[0][@{}l@{\AgdaIndent{0}}]%
\>[2]\AgdaFunction{calculateGenderAssignment}\AgdaSpace{}%
\AgdaBound{g}\AgdaSpace{}%
\AgdaBound{f}%
\>[33]\AgdaOperator{\AgdaFunction{≥}}\AgdaSpace{}%
\AgdaFunction{calculateLowerbound}\AgdaSpace{}%
\AgdaBound{g}\<%
\end{code}
We have defined a fairness property for a single gender we want to enrich an action handler so that applying an action is fair for all genders not just one. This is modelled by adding a \agdaFunction{IsFairForAll} type that is the product of the $IsFair$ type for all genders. 
\begin{code}
\>[0]\AgdaFunction{IsFairForAll}\AgdaSpace{}%
\AgdaSymbol{:}\AgdaSpace{}%
\AgdaFunction{TripCount}\AgdaSpace{}%
\AgdaSymbol{→}\AgdaSpace{}%
\AgdaPrimitiveType{Set}\<%
\\
\>[0]\AgdaFunction{IsFairForAll}\AgdaSpace{}%
\AgdaBound{f}\AgdaSpace{}%
\AgdaSymbol{=}\AgdaSpace{}%
\AgdaSymbol{∀}\AgdaSpace{}%
\AgdaSymbol{(}\AgdaBound{g}\AgdaSpace{}%
\AgdaSymbol{:}\AgdaSpace{}%
\AgdaDatatype{Gender}\AgdaSymbol{)}\AgdaSpace{}%
\AgdaSymbol{→}\AgdaSpace{}%
\AgdaFunction{IsFair}\AgdaSpace{}%
\AgdaBound{g}\AgdaSpace{}%
\AgdaBound{f}\<%
\end{code}
There are still two problems with implementing the action handler just using the $IsFairForAll$ type. The first problem is that it is unreasonable to assume that after the assignment of one or just a few trips that the trips will be fairly assigned. To model this we add the following predicate:
\begin{code}
\>[0]\AgdaFunction{UnderMinimumTripThreshold}\AgdaSpace{}%
\AgdaSymbol{:}\AgdaSpace{}%
\AgdaFunction{TripCount}\AgdaSpace{}%
\AgdaSymbol{→}\AgdaSpace{}%
\AgdaPrimitiveType{Set}\<%
\\
\>[0]\AgdaFunction{UnderMinimumTripThreshold}\AgdaSpace{}%
\AgdaBound{tripCount}\AgdaSpace{}%
\AgdaSymbol{=}\<%
\\
\>[0][@{}l@{\AgdaIndent{0}}]%
\>[2]\AgdaFunction{totalTripsTaken}\AgdaSpace{}%
\AgdaBound{tripCount}\AgdaSpace{}%
\AgdaOperator{\AgdaFunction{<}}\AgdaSpace{}%
\AgdaFunction{totalDrivers}\AgdaSpace{}%
\AgdaOperator{\AgdaPrimitive{*}}\AgdaSpace{}%
\AgdaNumber{10}\<%
\end{code}
The second problem is that there are two actions \agdaFunction{drive} and \agdaFunction{drivePassenger} and only the latter should count as a paying trip for the purpose of fairness. Again this is represented by another predicate:
\begin{code}
\>[0]\AgdaFunction{TripAgnostic}\AgdaSpace{}%
\AgdaSymbol{:}\AgdaSpace{}%
\AgdaDatatype{Action}\AgdaSpace{}%
\AgdaSymbol{→}\AgdaSpace{}%
\AgdaPrimitiveType{Set}\<%
\\
\>[0]\AgdaFunction{TripAgnostic}\AgdaSpace{}%
\AgdaSymbol{(}\AgdaInductiveConstructor{drivePassenger}\AgdaSpace{}%
\AgdaBound{t}\AgdaSpace{}%
\AgdaBound{p1}\AgdaSpace{}%
\AgdaBound{l1}\AgdaSpace{}%
\AgdaBound{l2}\AgdaSymbol{)}\AgdaSpace{}%
\AgdaSymbol{=}\AgdaSpace{}%
\AgdaDatatype{⊥}\<%
\\
\>[0]\AgdaFunction{TripAgnostic}\AgdaSpace{}%
\AgdaSymbol{(}\AgdaInductiveConstructor{drive}\AgdaSpace{}%
\AgdaBound{t}\AgdaSpace{}%
\AgdaBound{l1}\AgdaSpace{}%
\AgdaBound{l2}\AgdaSymbol{)}\AgdaSpace{}%
\AgdaSymbol{=}\AgdaSpace{}%
\AgdaRecord{⊤}\<%
\end{code}
We now have sufficient definitions to describe the fairness property in detail, in which an action is fair if it satisfies any of the three predicates defined above:
\begin{code}
\>[0]\AgdaKeyword{data}\AgdaSpace{}%
\AgdaDatatype{ActionPreservesFairness}\<%
\\
\>[0][@{}l@{\AgdaIndent{0}}]%
\>[2]\AgdaSymbol{(}\AgdaBound{α}\AgdaSpace{}%
\AgdaSymbol{:}\AgdaSpace{}%
\AgdaDatatype{Action}\AgdaSymbol{)}\AgdaSpace{}%
\AgdaSymbol{(}\AgdaBound{tripCount}\AgdaSpace{}%
\AgdaSymbol{:}\AgdaSpace{}%
\AgdaFunction{TripCount}\AgdaSymbol{)}\AgdaSpace{}%
\AgdaSymbol{:}\AgdaSpace{}%
\AgdaPrimitiveType{Set}\AgdaSpace{}%
\AgdaKeyword{where}\<%
\\
\>[2]\AgdaInductiveConstructor{underThreshold}\AgdaSpace{}%
\AgdaSymbol{:}\AgdaSpace{}%
\AgdaFunction{UnderMinimumTripThreshold}\AgdaSpace{}%
\AgdaBound{tripCount}\<%
\\
\>[2][@{}l@{\AgdaIndent{0}}]%
\>[4]\AgdaSymbol{→}\AgdaSpace{}%
\AgdaDatatype{ActionPreservesFairness}\AgdaSpace{}%
\AgdaBound{α}\AgdaSpace{}%
\AgdaBound{tripCount}\<%
\\
\>[2]\AgdaInductiveConstructor{fairForAll}\AgdaSpace{}%
\AgdaSymbol{:}\AgdaSpace{}%
\AgdaFunction{IsFairForAll}\AgdaSpace{}%
\AgdaBound{tripCount}\<%
\\
\>[2][@{}l@{\AgdaIndent{0}}]%
\>[4]\AgdaSymbol{→}\AgdaSpace{}%
\AgdaDatatype{ActionPreservesFairness}\AgdaSpace{}%
\AgdaBound{α}\AgdaSpace{}%
\AgdaBound{tripCount}\<%
\\
\>[2]\AgdaInductiveConstructor{agnostic}\AgdaSpace{}%
\AgdaSymbol{:}\AgdaSpace{}%
\AgdaFunction{TripAgnostic}\AgdaSpace{}%
\AgdaBound{α}\<%
\\
\>[2][@{}l@{\AgdaIndent{0}}]%
\>[4]\AgdaSymbol{→}\AgdaSpace{}%
\AgdaDatatype{ActionPreservesFairness}\AgdaSpace{}%
\AgdaBound{α}\AgdaSpace{}%
\AgdaBound{tripCount}\<%
\end{code}
The type of enriched action handlers that enforce this property can then be defined as follows:
\begin{code}
\>[0]\AgdaFunction{GenderAwareActionHandler}\AgdaSpace{}%
\AgdaSymbol{:}\AgdaSpace{}%
\AgdaPrimitiveType{Set}\<%
\\
\>[0]\AgdaFunction{GenderAwareActionHandler}\AgdaSpace{}%
\AgdaSymbol{=}\<%
\\
\>[0][@{}l@{\AgdaIndent{0}}]%
\>[2]\AgdaSymbol{(}\AgdaBound{α}\AgdaSpace{}%
\AgdaSymbol{:}\AgdaSpace{}%
\AgdaDatatype{Action}\AgdaSymbol{)}\<%
\\
\>[2]\AgdaSymbol{→}\AgdaSpace{}%
\AgdaSymbol{\{}\AgdaBound{tripCount}\AgdaSpace{}%
\AgdaSymbol{:}\AgdaSpace{}%
\AgdaDatatype{TripCount}\AgdaSymbol{\}}\<%
\\
\>[2]\AgdaSymbol{→}\AgdaSpace{}%
\AgdaSymbol{\{}\AgdaBound{fair}\AgdaSpace{}%
\AgdaSymbol{:}\AgdaSpace{}%
\AgdaDatatype{ActionPreservesFairness}\AgdaSpace{}%
\AgdaBound{α}\AgdaSpace{}%
\AgdaBound{tripCount}\AgdaSymbol{\}}\<%
\\
\>[2]\AgdaSymbol{→}\AgdaSpace{}%
\AgdaFunction{World}\AgdaSpace{}%
\AgdaSymbol{→}\AgdaSpace{}%
\AgdaFunction{World}\<%
\end{code}
One thing to note is that the form of this definition is slightly different from that of the \AgdaFunction{FuelAwareActionHandler} defined in the previous section. Instead of adding \AgdaFunction{TripCount} as a part of a product with the \AgdaFunction{World}, we add it as an implicit argument. This is because, unlike fuel, we've chosen not to enforce any type-level relationships between the trip count before an after applying the action. Instead we will rely on our enriched execute function to update the trip count correctly. The disadvantage of this approach is that one cannot enforce relationships between actions and the additional enriched state at the type-level, however the advantage of this is that it allows us to use exactly the same form for the enriched handler and canonical handler instances:
\begin{code}
\>[0]\AgdaFunction{enriched-σ}\AgdaSpace{}%
\AgdaSymbol{:}\AgdaSpace{}%
\AgdaFunction{Γ}\AgdaSpace{}%
\AgdaSymbol{→}\AgdaSpace{}%
\AgdaFunction{GenderAwareActionHandler}\<%
\\
\>[0]\AgdaFunction{enriched-σ}\AgdaSpace{}%
\AgdaBound{Γ}\AgdaSpace{}%
\AgdaBound{α}\AgdaSpace{}%
\AgdaSymbol{=}\AgdaSpace{}%
\AgdaFunction{updateWorld}\AgdaSpace{}%
\AgdaSymbol{(}\AgdaFunction{effects}\AgdaSpace{}%
\AgdaSymbol{(}\AgdaBound{Γ}\AgdaSpace{}%
\AgdaBound{α}\AgdaSymbol{))}\<%
\end{code}
Another advantage of working in a rich dependently-type language such as Agda is that our execution function can return error messages containing proofs in them explaining exactly why the execution of the function failed. Currently a failed execution just returns a precise error for why an execution has failed however we envision that we could use these precise errors for plan repair in future work. In this case our error contains a proof of why the action is not fair:
\begin{code}
\>[0]\AgdaKeyword{data}\AgdaSpace{}%
\AgdaDatatype{GenderBiasError}\AgdaSpace{}%
\AgdaSymbol{:}\AgdaSpace{}%
\AgdaPrimitiveType{Set}\AgdaSpace{}%
\AgdaKeyword{where}\<%
\\
\>[0][@{}l@{\AgdaIndent{0}}]%
\>[2]\AgdaInductiveConstructor{notProportional}\AgdaSpace{}%
\AgdaSymbol{:}\AgdaSpace{}%
\AgdaSymbol{(}\AgdaBound{a}\AgdaSpace{}%
\AgdaSymbol{:}\AgdaSpace{}%
\AgdaDatatype{Action}\AgdaSymbol{)}\AgdaSpace{}%
\AgdaSymbol{(}\AgdaBound{f}\AgdaSpace{}%
\AgdaSymbol{:}\AgdaSpace{}%
\AgdaFunction{TripCount}\AgdaSymbol{)}\<%
\\
\>[2][@{}l@{\AgdaIndent{0}}]%
\>[4]\AgdaSymbol{→}\AgdaSpace{}%
\AgdaOperator{\AgdaFunction{¬}}\AgdaSpace{}%
\AgdaSymbol{(}\AgdaDatatype{ActionPreservesFairness}\AgdaSpace{}%
\AgdaBound{a}\AgdaSpace{}%
\AgdaBound{f}\AgdaSymbol{)}\AgdaSpace{}%
\AgdaSymbol{→}\AgdaSpace{}%
\AgdaDatatype{GenderBiasError}\<%
\end{code}
The enriched execute function can be then be defined to check for fairness and can only execute in action if it can generate a proof that the action will not result in any gender bias:
\begin{code}
\>[0]\AgdaFunction{execute'}\AgdaSpace{}%
\AgdaSymbol{:}%
\>[747I]\AgdaDatatype{Plan}\AgdaSpace{}%
\AgdaSymbol{→}\<%
\\
\>[.][@{}l@{}]\<[747I]%
\>[11]\AgdaFunction{GenderAwareActionHandler}\AgdaSpace{}%
\AgdaSymbol{→}\<%
\\
\>[11]\AgdaFunction{TripCount}\AgdaSpace{}%
\AgdaSymbol{→}\<%
\\
\>[11]\AgdaFunction{World}\AgdaSpace{}%
\AgdaSymbol{→}\<%
\\
\>[11]\AgdaFunction{World}\AgdaSpace{}%
\AgdaOperator{\AgdaDatatype{⊎}}\AgdaSpace{}%
\AgdaDatatype{GenderBiasError}\<%
\\
\>[0]\AgdaFunction{execute'}\AgdaSpace{}%
\AgdaInductiveConstructor{halt}%
\>[17]\AgdaBound{σ}\AgdaSpace{}%
\AgdaBound{tripCount}\AgdaSpace{}%
\AgdaBound{w}\AgdaSpace{}%
\AgdaSymbol{=}\AgdaSpace{}%
\AgdaInductiveConstructor{inj₁}\AgdaSpace{}%
\AgdaBound{w}\<%
\\
\>[0]\AgdaFunction{execute'}\AgdaSpace{}%
\AgdaSymbol{(}\AgdaBound{a}\AgdaSpace{}%
\AgdaOperator{\AgdaInductiveConstructor{∷}}\AgdaSpace{}%
\AgdaBound{f}\AgdaSymbol{)}\AgdaSpace{}%
\>[17]\AgdaBound{σ}\AgdaSpace{}%
\AgdaBound{tripCount}\AgdaSpace{}%
\AgdaBound{w}\AgdaSpace{}%
\AgdaKeyword{with}\AgdaSpace{}%
\AgdaFunction{updateTripCount}\AgdaSpace{}%
\AgdaBound{a}\AgdaSpace{}%
\AgdaBound{tripCount}\<%
\\
\>[0]\AgdaSymbol{...}\AgdaSpace{}%
\>[6]\AgdaSymbol{|}\AgdaSpace{}%
\AgdaBound{updatedTrips}\AgdaSpace{}%
\AgdaKeyword{with}\AgdaSpace{}%
\AgdaFunction{ActionPreservesFairness?}\AgdaSpace{}%
\AgdaBound{a}\AgdaSpace{}%
\AgdaBound{updatedTrips}\<%
\\
\>[0]\AgdaSymbol{...}\AgdaSpace{}\AgdaSpace{}\AgdaSpace{}%
\>[8]\AgdaSymbol{|}\AgdaSpace{}%
\AgdaInductiveConstructor{yes}\AgdaSpace{}%
\AgdaBound{fair}\AgdaSpace{}%
\AgdaSymbol{=}\AgdaSpace{}%
\AgdaFunction{execute'}\AgdaSpace{}%
\AgdaBound{f}\AgdaSpace{}%
\AgdaBound{σ}\AgdaSpace{}%
\AgdaBound{updatedTrips}\AgdaSpace{}%
\AgdaSymbol{(}\AgdaBound{σ}\AgdaSpace{}%
\AgdaBound{a}\AgdaSpace{}%
\AgdaSymbol{\{}\AgdaArgument{fair}\AgdaSpace{}%
\AgdaSymbol{=}\AgdaSpace{}%
\AgdaBound{fair}\AgdaSymbol{\}}\AgdaSpace{}%
\AgdaBound{w}\AgdaSymbol{)}\<%
\\
\>[0]\AgdaSymbol{...}\AgdaSpace{}\AgdaSpace{}\AgdaSpace{}%
\>[8]\AgdaSymbol{|}\AgdaSpace{}%
\AgdaInductiveConstructor{no}\AgdaSpace{}%
\AgdaBound{¬fair}\AgdaSpace{}%
\AgdaSymbol{=}\AgdaSpace{}%
\AgdaInductiveConstructor{inj₂}\AgdaSpace{}%
\AgdaSymbol{(}\AgdaInductiveConstructor{notProportional}\AgdaSpace{}%
\AgdaBound{a}\AgdaSpace{}%
\AgdaBound{updatedTrips}\AgdaSpace{}%
\AgdaBound{¬fair}\AgdaSymbol{)}\<%
\end{code}

\section{Implementation, Code Extraction, Further Applications}
\label{sec:applications}

The accompanying Agda library~\cite{DFHKS21} is arranged in a way that is friendly to users from the AI planning community. This is the summary of the general methodology to set up, verify and execute a PDDL problem using an enriched handler in our Agda library: 


\begin{enumerate}
\item Import the \textsf{Semantics} and \textsf{Plan} files from the Plan folder.
\item Create and import a \textsf{Domain} file for your problem. 
\item Define an initial world, goal state and a plan.
\item Use the typing derivation to check that the plan is valid for the initial world and goal state provided.
\item Create an enriched handler and evaluation function:
\begin{enumerate}
\item Model the additional properties as types.
\item Show that the additional properties are decidable if necessary.
\item Create the relevant error types. 
\item Define an action handler that includes the additional properties.
\item Define an evaluation function for the action handler.
\item Define an enriched canonical handler.
\end{enumerate}
\item Import the enriched handler that you want to use.
\item Use the relevant evaluation function to execute your handler on the initial world.
\end{enumerate} 

Although the primary purpose of the presented work is to test the limits of type-driven code development in AI, we have put some thought into future extensions and applications of this work.

In this paper we have manually performed steps 2,~3~and~4, however in previous work ~\cite{HillKP20} these steps were fully automated to relieve the burden on the user. The automation is not immediately transferable to the paper due to changes in the Agda formalisation mentioned in Section~\ref{sec:ah}, however it would be relatively easy to update it.

As for applications to AI planning, we envisage several. Firstly,
one can use our implementation of the plan validator to verify plans using the typing relation. In ~\cite{HillKP20} we showed a few examples of when this exercise can reveal surprising (and often undesirable)
properties of plans produced by STRIPS.

A second use for this methodology is suggested by Agda's infrastructure for code extraction. It is easy enough to extract the examples that we implemented either to Haskell or to binaries, with the repository~\cite{DFHKS21} contains some detailed description of the extracted files we obtain as a result (accompanying previous papers on this topic). Thus, one can imagine future deployment of such verified code directly on to robots.  

Finally, and as discussed, there may be use cases when software and hardware requirements, or indeed legal regulations, do not permit the direct deployment of
code extracted from Agda. For example in the autonomous car industry, the set of usable languages is strictly regulated.
In such cases, the methodology we proposed can be used as part of a broader modelling and simulation environment.
In fact, we believe this to be the most promising avenue for applications of these ideas.  The enriched handlers proposed in this paper enhance exactly this modelling aspect, by opening a way for lightweight and flexible
modelling of arbitrary properties of plans separately from (and in addition to) the automated plan search performed by an AI planner such as STRIPS. 

%
%



\section{Conclusions, Related and Future Work}\label{sec:future}
\label{sec:relatedwork}

We have presented a novel methodology of using enriched handlers for embedding AI plans into a richer programming and modelling environment in Agda.
Our main focus was to show that the idea of a verification framework combining automated solvers and planners on the one hand and
richer type-driven programming environments on the other hand has its merits, and can be implemented in an interesting, natural and even user-friendly way. 
We hope that this line of work inspires more applications in AI verification in the future. 

Apart from our own work~\cite{HillKP20,SchwaabKHFPWH19}, we are not aware of any other approaches to (dependent) type driven methodology for AI plans.
However, more broadly the logic and programming language communities have paid attention to AI planning in the past.

\textbf{AI Planning and Linear Logic.}
There is a long history of modelling AI planning in Linear logic, that dates back to the 90s~\cite{jacopin1993classical}, and was investigated in detail in the 2000s, see e.g.
\cite{chrpa2007encoding,steedman2002plans}.
In fact, AI planning is used as one of the iconic use-cases of Linear logic~\cite{polakow2001ordered}.
The main idea behind using Linear logic for AI planning is
treating action descriptions  as linear implications:
$$\alpha : \forall x. P \multimap Q,$$
where  $P$ and $Q$ are given by tensor products of atoms:
$R_1(t_1) \otimes \ldots \otimes R_n(t_n)$.
We could incorporate information about polarities inside the predicates, as follows:
$R_1(t_1,z_1) \otimes \ldots \otimes R_n(t_n,z_n)$.
Then, the linear implication and the tensor products model the resource semantics of PDDL rather elegantly.

The computational (Curry-Howard) interpretation of AI plans was not the focus of study in the above mentioned approaches,
yet it plays a crucial role in this paper, from design all the way to implementation, verification and proof extraction.

\textbf{AI Planning and (Linear) Logic Programming.}
The above syntax also resembles linear logic programming Lolli, introduced by Miller et al~\cite{HodasM94}.
Lolli was applied in speech planning in~\cite{DixonST09}.

Our previous work~\cite{SchwaabKHFPWH19} in fact takes inspiration from Curry-Howard interpretation of Prolog~\cite{0001KSP16,FuK17}.
In our previous work and in general, logic programming does not work well with PDDL negation. In PDDL, we have to work with essentially three-valued logic:
a predicate may be declared to be absent or present in a world. But if neither is declared, we assume a ``not known'' or ``either'' situation.
Logic programming usually uses the approach of ``negation-as-failure'' that does not agree with this three-valued semantics.
A solution is to introduce polarities as terms, as shown in the example above. This merits further investigation.

\textbf{Curry-Howard view on Linear Logic.}
Curry-Howard semantics of Linear logic also attracted attention of logicians first in the 90s~\cite{AlbrechtCJ97}, and then in the 2000s
in connection with research into Linear Logical Frameworks~\cite{Schack-NielsenS08,CervesatoP02}.

The work that we do relates to that line of work, and can be seen as a DSL for AI planning. It is simpler and
less expressive than Linear logic generally but makes up for it in simplicity and close correspondence
to PDDL syntax. Transformations between PDDL domain and problem descriptions to Agda syntax
are straightforward by design of the DSL.
This enables us to automate the generation of Agda proofs from PDDL plans.

\textbf{Origins of the Frame Rule.}
The ``frame problem'' that inspired the frame rule of Separation logic
actually
has origins in AI~\cite{hayes1981frame,dennett2006cognitive}.
Initially, the problem referred to the difficulty in local reasoning about problems in a complex
world. In AI planning specifically, this problem consisted of keeping track
of the consequences of applying an action on a world. Intuitively, one understands
that driving one passenger in one taxi would have no effect on a journey time of another passenger in another taxi. 
The frame problem deals with the
way to represent this intuition formally.

One way to deal with the frame problem is to declare ``frame axioms'' for every action
explicitly. This is an inefficient way to deal with this
problem as defining these frame axioms becomes infeasible the larger the system
gets~\cite{dennett2006cognitive}. Since most actions in AI planning only make small
local changes to the world, a more general representation would be more suitable.
STRIPS deals with this problem by introducing an assumption that every formula
in a world that is not mentioned in the effect list of an action remains the same
after execution of the action. This is known as the ``STRIPS assumption'' and
it is an assumption that PDDL also uses.

The logic of Bunched Implications \cite{o2001local,ishtiaq2001bi} and Separation Logic~\cite{o2007resources}
took inspiration from this older notion of the frame problem, and introduced more abstract formalism,
which is now known as a ``frame rule'', into the resource logics~\cite{pym2019resource}.
This family of logics has brought many theoretical and practical advances to modelling of complex systems, and is behind many \emph{lightweight verification}
projects~\cite{calcagno2015moving}.

Outside of logic and semantics communities, AI planning researchers recently started to invest more effort into explaing and validating plans, as well as in modelling extrinsic properties.  We highlight two approaches in particular. 

\textbf{Formalisation of planning in other theorem provers.}
Much of the work we proposed here could be replicated in another dependently-typed prover, such as Coq, Idris or Lean.
In addition, there has been an impressive line of work by Abdulaziz and co-authors on formal verification of plan validators in Isabelle/HOL, see e.g.
\cite{abdulaziz2018formally,AbdulazizB21,AbdulazizL20}. One advantage of using Isabelle/HOL for PDDL  formalisations is availability of extensive mathematical libraries that can support proof development.
On the other hand, dependently-typed provers are particularly attractive for ``lightweight verification'' via type-driven program development. This is exactly the feature we wanted to show-case in this paper.

\textbf{Explainable AI.}
Extrinsic tools that introduce meta properties over PDDL are already being used in the field of Explainable AI. In \cite{cashmore2019towards} a wrapper over PDDL was created so that users can express ``contrastive questions'' to better understand and explore why a planner has chosen certain actions over others. An example contrastive question could be \emph{"Why did you choose action A rather than B?"}. To accomplish this, users give questions in natural language which are then converted into formal constraints that are then compiled down into PDDL. These additional constraints force the planner to choose different actions which the wrapper will use to generate a contrastive explanation by comparing the original plan to the new plan generated from the additional constraints. The user can then add additional constraints by asking further contrastive questions. This ability to ask further questions is particularly useful as it allows a user to build complex constraints to gain a deeper understanding of a plan. 

\textbf{Plan-property Dependencies.}
There is also work \cite{eifler2020new,eifler2020plan} that introduces plan-property dependencies which impose boolean functions over plans which allows a user to query why a plan satisfies certain properties over others. These properties are equivalent to soft goals in PDDL~\cite{gerevini2005plan}. This work explains plans by showing the cost of satisfying certain properties over others by computing the minimal unsolvable goal subsets of a planning problem. An example question in this work could be \emph{"Why does the plan not satisfy the property X?"} and a potential reply could be \emph{"because then we would have to forgo property Y and property Z"}. To be able to do this, they compile plan properties into goal facts and then compute the minimal unsolvable goal subsets to produce plan explanations. This work can also reason about plan properties in linear temporal logic. 

In comparison to our work, both of the previous approaches define extrinsic properties in a domain-independent manner. Whilst the verification and execution of plans in our system is domain-independent, the enriched handlers are not necessarily domain independent. For example, the more generic properties of \agdaFunction{FuelAwareActionHandler} could be used in any domain, however the \agdaFunction{GenderAwareActionHandler} is defined specifically for the taxi domain. The benefit of our approach is that we can define complex properties that would be undefinable in either of the previous systems. However, at the current moment we have no way to compile our properties into PDDL when a plan fails.

One area of future work that we would like to focus on is plan repair. In our current system we can verify additional properties of plans using our enriched handlers but we have no obvious course of action for what to do once a plan fails. In this paper we have tried to address this by choosing additional constraints that will most likely be satisfied by a planner without any additional replanning. We believe that we could address this issue by compiling down additional constraints to PDDL based on the extrinsic properties of the enriched handler. Since the extrinsic properties
can not be easily expressed in PDDL we can create compilation strategies based on the errors produced by failed evaluations to force the planner to pick different actions. For example, if we have a plan that fails in our taxi domain because it has disproportionately picked men over women in a plan we could fix this by removing a certain number of male taxi drivers from the planning problem so that the planner no longer has the option to choose them. This could be further enhanced by modelling partial plans where a new PDDL problem can be created at a failure point in a plan. This would potentially reduce the amount of replanning needed. 

In previous work~\cite{HillKP20} we fully automated our system so that verification and execution of plans can be generated from PDDL domain and problem descriptions. This should ensure that there is a low barrier for entry for new users in terms of Agda and programming language knowledge.
Because the extrinsic properties (modelled by the enriched handlers) are not part of the PDDL domain or problem,
we cannot provide the same level of automation for generating these.
In future work we intend to address this by creating a more user-friendly infrastructure for defining the extrinsic properties.
For example, a DSL for enriched handlers is an option worth considering.
This would mean that a user would only have to learn how to use the DSL.
Implementing such a DSL may even open opportunities for  automating the feedback loop from Agda to PDDL.
A drawback of this approach would be that we will have to restrict the expressibility of enriched handlers.


\begin{acks}
  The first author acknowledges support of the EPSRC Doctoral Training scheme; the second and third author acknowledge
  generous support of the EPSRC grant \emph{AISEC: AI Secure and Explainable by Construction}, EP/T026952/1, \url{https://www.macs.hw.ac.uk/aisec/}.  The authors also thank the TyDe'21 reviewers for their valuable input, which lead to substantial clarification and simplifaction of some of the code
  accompanying this paper.
 
\end{acks}

\balance
\bibliographystyle{ACM-Reference-Format}
\bibliography{mybib}

\end{document}